\documentclass[arxiv]{melba}

\usepackage{mwe} 

\usepackage{amsmath,amsfonts}

\usepackage{tikz}
\usepackage{subfig} 

\usepackage{rotating} %
\usepackage{xcolor,colortbl}
\usepackage{graphicx}

\DeclareMathOperator{\DSC}{DSC}
\DeclareMathOperator{\ASSD}{ASSD}

\definecolor{Gray}{gray}{0.85}
\definecolor{LightGray}{gray}{0.95}
\definecolor{Skeleton Suns-c}{rgb}{0.86,0.371,0.34}
\definecolor{IMR-c}{rgb}{0.829,0.86,0.34}
\definecolor{CompAI-c}{rgb}{0.34,0.86,0.371}
\definecolor{Hilab-c}{rgb}{0.34,0.829,0.86}
\definecolor{IMI-c}{rgb}{0.371,0.34,0.86}
\definecolor{SKJP-c}{rgb}{0.86,0.34,0.829}

\DeclareRobustCommand\circle[1]{\setlength{\fboxrule}{0pt}%
\fbox{\tikz\draw[#1,fill=#1] (0,0) circle (.7ex);}}

\melbaid{2025:001}  
\doi{10.59275/j.melba.2025-d482}
\melbaauthors{Dorent, \textit{et al.}}  
\email{reuben.dorent@inria.fr}
\volume{2}
\firstpageno{1}  
\melbayear{2025}  
\datesubmitted{2024-08-20}  
\datepublished{2025-01-29}  

\melbaspecialissue{\textbf{MICCAI 2023 LNQ challenge special issue}}
\melbaspecialissueeditors{Steve Pieper, Erik Ziegler, Tawa Idris, Bhanusupriya Somarouthu, Reuben Dorent, Gordon Harris, Ron Kikinis}

\ShortHeadings{LNQ 2023 challenge: Benchmark of weakly-supervised techniques for mediastinal lymph node quantification}{Dorent, \textit{et al.}}

\title{LNQ 2023 challenge: Benchmark of weakly-supervised techniques for mediastinal lymph node quantification}

\author{
	\firstname Reuben \surname Dorent\aff{1}\orcid{0000-0002-7530-0644}, 
    \firstname Roya \surname Khajavi\aff{1}, 
    \firstname Tagwa \surname Idris\aff{2}\orcid{0000-0003-4524-7974},  
    \firstname Erik \surname Ziegler\aff{3}\orcid{0000-0003-1857-8129},
    \firstname  Bhanusupriya \surname Somarouthu\aff{2,3}\orcid{0009-0007-7081-5423},
    \firstname Heather \surname Jacene\aff{1,5}\orcid{0000-0001-7167-078X},
    \firstname Ann \surname LaCasce\aff{5},
    \firstname Jonathan \surname Deissler\aff{16}
    \firstname Jan \surname Ehrhardt\aff{10,11},
    \firstname Sofija \surname Engelson\aff{10}\orcid{0009-0007-2493-8107},
    \firstname Stefan M. \surname Fischer\aff{6,7,8}\orcid{0009-0005-7637-7183},
    \firstname Yun \surname Gu\aff{14}\orcid{ 0000-0002-4199-0675},
    \firstname Heinz \surname Handels\aff{10,11}\orcid{0000-0002-3499-4328},
    \firstname Satoshi \surname Kasai\aff{12}\orcid{0009-0004-0067-1442},
    \firstname Satoshi \surname Kondo\aff{13},
    \firstname Klaus \surname Maier-Hein\aff{16,17}\orcid{0000-0002-6626-2463},
    \firstname Julia A. \surname Schnabel\aff{6,7,8,9}\orcid{0000-0001-6755-8672},
    \firstname Guotai \surname Wang\aff{15,18}\orcid{0000-0002-8632-158X}, 
    \firstname Litingyu \surname Wang\aff{15},
    \firstname Tassilo \surname Wald\aff{16,17}\orcid{0009-0007-5222-2683}, 
    \firstname Guang-Zhong \surname Yang\aff{14}\orcid{0000-0003-4060-4020}, 
    \firstname Hanxiao \surname Zhang\aff{14}\orcid{0000-0002-2671-8606},
    \firstname Minghui \surname  Zhang\aff{14}, 
    \firstname Steve \surname Pieper\aff{4}\orcid{0000-0003-4193-9578}, 
    \firstname Gordon \surname Harris\aff{2,3}\orcid{0000-0002-8664-7707}, 
    \firstname Ron \surname Kikinis\aff{1}\orcid{ 0000-0001-7227-7058}, 
    \firstname Tina \surname Kapur\aff{1}\orcid{0000-0003-3646-9508}
}
\affiliations{
	\num 1 \addr Brigham and Women’s Hospital, Harvard Medical School, Boston, MA, USA;
	\num 2 \addr Massachusetts General Hospital, Harvard Medical School, Boston, MA, USA; 
	\num 3 \addr Yunu, Inc., Cary, NC, USA;
    \num 4 \addr Isomics Inc, Cambridge, MA, USA;
    \num 5 \addr Dana-Farber Cancer Institute, Boston, MA, USA;
    \num 6 \addr Technical University Munich, Munich, Germany;
    \num 7 \addr Institute of Machine Learning in Biomedical Imaging, Helmholtz Munich, Munich,  Germany;
    \num 8 \addr  Munich Center of Machine Learning (MCML), Munich, Germany;
    \num 9 \addr School of Biomedical Engineering and Imaging Sciences, King's College London, London, UK;
    \num 10 \addr Institute of Medical Informatics, University of Lübeck, Lübeck, Germany;
    \num 11 \addr German Research Center for Artificial Intelligence, Lübeck, Germany;
    \num 12 \addr Niigata University of Health and Welfare, Niigata, Japan;
    \num 13 \addr Muroran Institute of Technology, Hokkaido, Japan;
    \num 14 \addr Institute of Medical Robotics, Shanghai Jiao Tong University, Shanghai, China;
    \num 15 \addr University of Electronic Science and Technology of China, Chengdu, China;
    \num 16 \addr Division of Medical Image Computing, German Cancer Research Center (DKFZ), Heidelberg, Germany;
    \num 17 \addr University of Heidelberg, Heidelberg, Germany;
    \num 18 \addr Shanghai AI laboratory, Shanghai, China.
}

\abstract{
Accurate assessment of lymph node size in 3D CT scans is crucial for cancer staging, therapeutic management, and monitoring treatment response. Existing state-of-the-art segmentation frameworks in medical imaging often rely on fully annotated datasets. However, for lymph node segmentation, these datasets are typically small due to the extensive time and expertise required to annotate the numerous lymph nodes in 3D CT scans. Weakly-supervised learning, which leverages incomplete or noisy annotations, has recently gained interest in the medical imaging community as a potential solution. Despite the variety of weakly-supervised techniques proposed, most have been validated only on private datasets or small publicly available datasets.
To address this limitation, the Mediastinal Lymph Node Quantification (LNQ) challenge was organized in conjunction with the 26th International Conference on Medical Image Computing and Computer Assisted Intervention (MICCAI 2023). This challenge aimed to advance weakly-supervised segmentation methods by providing a new, partially annotated dataset and a robust evaluation framework. A total of 16 teams from 5 countries submitted predictions to the validation leaderboard, and 6 teams from 3 countries participated in the evaluation phase.
The results highlighted both the potential and the current limitations of weakly-supervised approaches. On one hand, weakly-supervised approaches obtained relatively good performance with a median Dice score of $61.0\%$. On the other hand, top-ranked teams, with a median Dice score exceeding $70\%$, boosted their performance by leveraging smaller but fully annotated datasets to combine weak supervision and full supervision. This highlights both the promise of weakly-supervised methods and the ongoing need for high-quality, fully annotated data to achieve higher segmentation performance.
}

\keywords{Machine Learning, Image Segmentation, Weak Supervision, Lymph Node}

\begin{document}

\twocolumn[\maketitle]

\section{Introduction}\enluminure{M}{achine} learning has recently achieved outstanding medical image segmentation performance. 
\sloppy While initial frameworks were tailored to specific segmentation tasks, significant progress has been made in developing network architectures and training procedures that are robust and applicable across segmentation tasks~\citep{isensee2021nnu,cardoso2022monai}. These robust frameworks generally rely on the availability of large, fully annotated datasets during training. This reliance, however, is often impratical. Manual medical data segmentation is a labor-intensive and time-consuming process that demands rare expertise, leading to small, fully annotated datasets. Moreover, obtaining complete annotations in an image is sometimes unfeasible due to factors such as the disproportionate size of the targeted object of interest compared to the image (e.g., cells in histopathology images), data ambiguity (e.g., tumor in ultrasound imaging), or the presence of a very large number of objects to be segmented (e.g., lymph nodes in 3D scans).

In response to these challenges, weakly-supervised learning methods have emerged as promising alternatives. These approaches leverage sparse, noisy, or incomplete annotations to train machine learning models, significantly reducing the need for complete annotations at training time. In the context of medical imaging, weakly-supervised learning can utilize various forms of weak supervision, such as image-level labels \citep{ouyang2019weakly}, bounding boxes \citep{pmlr-v121-kervadec20a}, 
scribbles \citep{zhang2022cyclemix, dorent2020scribble}, points \citep{roth2019weakly,dorent2021inter, can2018learning}, 
linear measurements \citep{cai2018accurate, li2020deep} or partial annotations \citep{mehrtash2024evaluation, dorent2021learning}, to achieve competitive performance in segmentation tasks.

To benchmark new and existing weakly-supervised techniques for medical image segmentation, we organized the Mediastinal Lymph Node Quantification (LNQ) challenge in conjunction with the 26th International Conference on Medical Image Computing and Computer Assisted Intervention (MICCAI 2023). The challenge's goal was to segment and identify mediastinal lymph nodes in contrast-enhanced computed tomography (CT) scans using a new large and partially annotated dataset. Note that participants were allowed to exploit existing smaller, publicly available, and fully annotated datasets.

Accurate lymph node size estimation is critical for staging cancer patients, initial therapeutic management, and, in longitudinal scans, assessing response to therapy. Current standard practice for quantifying lymph node size is based on various criteria that use unidirectional or bidirectional measurements on just one or a few nodes, typically on just one axial slice. These evaluations are performed on routine CT scans. However, humans have hundreds of lymph nodes, any number of which may be enlarged to various degrees due to disease or immune response. While a normal lymph node may be approximately 5 mm in diameter, a diseased one may be several cm in diameter. The mediastinum, the anatomical area between the lungs and around the heart, may contain ten or more lymph nodes, often with three or more enlarged greater than 1 cm in afflicted patients. Accurate volumetric assessment would thus provide information to evaluate lymph node disease and provide better sensitivity to detect volumetric changes indicating response to therapy.

While automated full 3D segmentation of all abnormal lymph nodes could improve cancer treatment, only small, fully annotated datasets are currently publicly available to train machine learning frameworks. For these reasons, we proposed a weakly-supervised benchmark that aims to automatically perform lymph node segmentation in 3D CT scans using partial annotations. Specifically, we partially annotated (a few nodes only) a large number of scans and evaluated the performance of the participants' methods using a large, fully annotated dataset.

 This paper summarizes the LNQ 2023 challenge and is structured as follows. First, a review of existing datasets used to perform lymph node quantification is presented in Section 2. Then, the design of the LNQ challenge is given in Section 3. Section 4 presents the evaluation strategy of the challenge (metrics and ranking scheme). Participating methods are then described and compared in Section 5. Finally, Section 6 presents the results of the participating teams, and Section 7 provides a discussion and concludes the paper.

\begin{figure*}[tb!]
  \includegraphics[width=\linewidth]{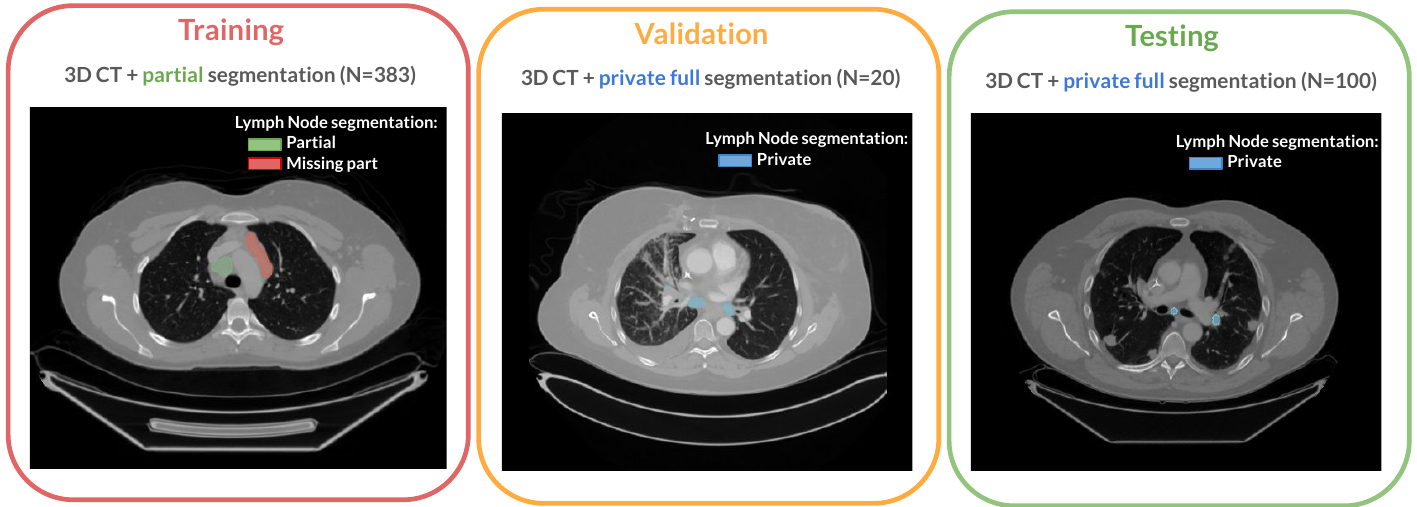}
  \caption{Overview of the challenge dataset. Only partial annotations (green) were made available for the training 3D CT scans. Missing training data nodes are shown in red. Full segmentations (blue) of all nodes were performed for evaluation on the validation and testing sets, and remained private.}
  \label{fig:data_set_pres}
\end{figure*} 

\section{Related Works}

We performed a literature review to survey the datasets used to assess lymph node quantification. Many lymph node quantification techniques have been explored and validated on private datasets, for example, \cite{tan2018, 6033061, STAPLEFORD2010959, feulner2013lymph}. Since these datasets are private and working implementations of the methods are not available, it is impossible to benchmark existing and new methods with these techniques. 

Other authors have used public datasets to validate their methods. We present these open datasets and highlight their limitations for evaluating lymph node quantification:
\begin{itemize}
    \item \textit{CT Lymph Node dataset}: The NIH CT Lymph Node dataset~\citep{roth2015lymphnodedata_seg} comprises a total of 176 contrast-enhanced CT series from 176 patients. Among them, 89 CT volumes were obtained at the chest level (mediastinum). Partial node segmentation is provided, corresponding to 387 nodes with a short axis diameter (SAD) $\geq$ 1 cm, which is considered clinically enlarged and abnormal.  \cite{bouget2023mediastinal} extended and refined these annotations for all the mediastinal lymph nodes in these 89 volumes. All suspicious regions are segmented as lymph nodes, including nodes with short-axis measurement less than 1 cm. In total, 2912 nodes were segmented.
    \item \textit{St. Olavs Hospital dataset}: This dataset~\citep{bouget2019semantic} comprises 15 contrast-enhanced CT volumes from 15 patients with confirmed lung cancer diagnoses. Segmentation of all the lymph nodes is provided. A total of 384 lymph nodes were annotated in this dataset.
 \end{itemize}

Moreover, other datasets that comprise contrast-enhanced
CT scans have been released from patients with lung cancer, such as NSCLC-Radiomics~\citep{aerts2014decoding}, NSCLC Radiogenomics~\citep{bakr2018radiogenomic}, NSCLC-Radiomics Interobserver1~\citep{wee2019data}, RIDER Lung CT~\citep{zhao2009evaluating}. However, these datasets do not provide any segmentations of lymph nodes.

In conclusion, test sets used to assess segmentation methods for mediastinal lymph node segmentation are either private, relatively small, or partially annotated.

On the methodological side, a wide range of weakly-supervised methods have been proposed.
Nonetheless, a thorough review of weakly-supervised methodologies is out of the scope of this paper, except to note that it it an active area of research with many unsolved challenges. We refer the interested reader to \cite{zhang2020survey} for a review of these.

\begin{table*}[tb]
	\centering
	\caption{Summary of characteristics of the LNQ data sets. Median [Q1-Q3] are given for numerical values.
	}\label{tab:DataCharacteristics}
	\resizebox{\textwidth}{!}{
	\begin{tabular}{l *{3}{c}}
		\toprule
		\multirow{1}{*}{} & \multicolumn{1}{c}{\bf Training} & \multicolumn{1}{c}{\bf Validation } & \multicolumn{1}{c}{\bf Test }\\ 
       \cmidrule(lr){2-2} \cmidrule(lr){3-3} \cmidrule(lr){4-4}
       
		Modality & contrast-enhanced CT & contrast-enhanced CT & contrast-enhanced CT \\ 
		\rowcolor{LightGray} 
        Number of scans  &  383 &  20  &  100  \\
        Number of patients  &  383 &  20  &  100   \\
        \rowcolor{LightGray}
         & Breast ($18\%$) & Breast ($25\%$) & Breast ($4\%$) \\
         \rowcolor{LightGray}
         & Chronic lymphocytic leukemia ($15\%$) & Chronic lymphocytic leukemia ($30\%$) & Chronic lymphocytic leukemia ($11\%$) \\
         \rowcolor{LightGray}
         & Hodgkin lymphoma ($7\%$) & Hodgkin lymphoma ($5\%$)& Hodgkin lymphoma ($8\%$)\\
         \rowcolor{LightGray}
         & Lung non-small cell ($9\%$) & Lung non-small cell ($25\%$) & Lung non-small cell ($16\%$)\\
         \rowcolor{LightGray}
         & Lung small cell ($6\%$) & Lung small cell  ($5\%$) & Lung small cell ($7\%$) \\
         \rowcolor{LightGray}
         & Renal cell ($5\%$) & Renal cell  ($0\%$) & Renal cell ($7\%$) \\
         \rowcolor{LightGray}
         \multirow{-7}{*}{Cancer type} & Others ($40\%$) & Others ($10\%$) & Others ($47\%$) \\
        Public annotations & Some lymph nodes & $\times$ & $\times$ \\
        \rowcolor{LightGray}
        Private annotations & $\times$ & All lymph nodes & All lymph nodes \\
        In-plane matrix & 512 ($100\%$) & 512 ($100\%$)& 512 ($100\%$)  \\
        \rowcolor{LightGray}
        Slice number  & 114 [96-136] & 112 [92-127] & 86 [77-93] \\
         In-plane res. in mm  & 0.8 [0.8-0.9] & 0.8 [0.8-0.9] & 0.9 [0.8-0.9] \\
        \rowcolor{LightGray}
        Slice spacing in mm  &  3.0 [2.5-3.8] & 3.0 [2.5-3.8] & 3.8 [3.8 - 5.0] \\
        Male:Female & $47\%:53\%$ & $45\%:55\%$ & $47\%:53\%$\\
		\bottomrule
	\end{tabular}
	}
\end{table*}
\newpage
\section{Challenge description}\label{sec:challenge_description}

\subsection{Overview}
The goal of the LNQ challenge was to benchmark new and existing weakly-supervised techniques for lymph node quantification. The proposed segmentation task focused on segmenting all the lymph nodes in contrast-enhanced computed tomography (CT) scans. Participants had access to a training set of partially annotated CT scans. Participant's algorithms were evaluated on a fully annotated dataset.

\subsection{Data description}
\subsubsection{Data overview}
The challenge cohort is a cross-institutional dataset of 513 chest CT scans acquired during patient treatment for various cancer types. The dataset originates from the Tumor Imaging Metrics Core (TIMC), a multi-institutional imaging core lab (Massachusetts General Hospital,
Dana-Farber Cancer Institute, Brigham and Women’s Hospital, Boston, MA, USA) that provides multimodality imaging measurements to evaluate treatment response in patients enrolled in oncology clinical trials. This dataset was used for training, validation, and testing. Figure~\ref{fig:data_set_pres} presents an overview of the challenge dataset.

The complete LNQ dataset (training, validation, and testing) contained CT images collected on 513 patients (Male:Female 239:274)  enrolled in oncology clinical trials from 2007-2020. For each patient, contrast-enhanced CT scans were acquired. The Institutional Review Board at the Mass General Brigham (MGB IRB) approved the protocol (2020P000211), including public sharing of the data (2020P003754).

The dataset comprises patients with various types of cancer. The top six primary patient cancers, which accounted for $60\%$ of all patients, were breast cancer (N=80), chronic lymphocytic leukemia (CLL) (N=74),  non-small cell lung cancer (N=58), Hodgkin‘s lymphoma (N=38), small cell lung cancer (N=30), and renal cell cancer (N=25). The remaining patient cancer types, accounting for $40\%$ of all patients, including thyroid cancer, adenocarcinoma, endometrial adenocarcinoma, melanoma, head and neck cancer, non-Hodgkin’s lymphoma, prostate cancer, mesothelioma, esophageal cancer, ovarian cancer, and colon cancer.

The radiology and oncology experts on our team have determined that for this challenge, the CT appearance of clinically important lymph nodes is not influenced by the primary cancer type. Based on this, we have included as many cases as possible with segmented lymph nodes of any primary cancer type on the premise that access to more data that experts judge to be similar in appearance is more likely to provide a robust segmentation model.

The dataset was randomly split into training ($76\%$), validation ($4\%$) and test ($20\%$) sets. Table~\ref{tab:DataCharacteristics} shows the distribution of features (patient sex, cancer, slice thickness, and in-plane resolution) across these sets.


\subsubsection{Image acquisition}

Images from the LNQ challenge were acquired with routine clinical CT scanners from various manufacturers (e.g. GE Healthcare Discovery CT750HD, GE Medical System BrightSpeed, Siemens SOMATOM Definition, Toshiba Aquilion, Philips iCT). The scans are of routine clinical quality, with 2 to 5mm slice spacing and an in-plane resolution of approximately $1\times1\text{mm}$ or smaller.  Acquisitions are typically 512x512 axial scans with 100 or more slices (range: 48-656 slices). 

\subsubsection{Annotation protocol}

All imaging datasets were manually segmented following two annotation protocols. While the training set was partially annotated, the validation and test sets were fully annotated, following the considerations described below.

The training set was annotated based on data from the clinical trials workflow. This annotation process involves two steps: lymph node selection and manual volumetric segmentation. Since the initial annotations are created within the context of clinical trials, the cases are not initially fully annotated; only specific lesions or lymph nodes are marked. Specifically, expert image analysts used the Yunu (Cary, NC, previously known as Precision Imaging Metrics (PIM)) clinical trials imaging informatics software system  to select lesions, focusing on those to be followed with bi-dimensional measurements in single axial slices of longitudinal CT scans. Nodes were selected according to trial protocols defined by the sponsor, which may vary for different patients. This means the selected lesions may not always be the largest present in the case \citep{eisenhauer2009new}. Moreover, protocols may have different selection criteria or only call for annotations of a maximum number of lesions per patient. In the second step, the volumetric extent of these selected nodes was manually segmented to create the weak annotations for the heterogeneous training CT data of the challenge.

The validation and test sets were fully segmented to reflect the full mediastinal lymph node disease burden and used to assess the performance of the LNQ participants' methods. All lymph nodes considered abnormal, with an estimated short axis length larger than 1 cm, were targeted for segmentation. Surrounding structures excluded from the lesion boundary included large vessels, artifacts, and non-nodal components. We acknowledge that visual assessment of the short axis is error-prone, so this criterion may not always be satisfied for nodes with a short axis around the threshold.

Volumetric segmentations were performed using 3D Slicer \citep{pieper20043d,fedorov20123d}. All lymph nodes were segmented by a trained radiologist (TI), with the support of an instructor in radiology (BS), and in consultation with a senior imaging specialist (HJ). For each patient, the CT scan and the bidimensional measurements of the target lymph node were loaded into 3D Slicer (version 4.11). The Segment Editor module was then used to manually delineate the lymph node boundary at the native resolution of the CT scan. The Draw tool within the Editor module was employed to draw freehand boundaries on axial cross-sections while referencing the sagittal and coronal planes.

\subsubsection{Data curation}
Data were fully de-identified by removing all personal health information identifiers and creating fresh DICOM files containing only approved tags and images. The MGB IRB (2020P000211 and 2020P003754), approved the de-identification procedure, and all header data was manually reviewed by MGB staff.
Image series were manually reviewed for quality and to ensure no issues were caused by de-identification. 

Images and segmentation masks were distributed as NRRD files (.nrrd). The training and validation datasets were made publicly available via the challenge page \footnote{\url{https://lnq2023.grand-challenge.org/data/}}. In contrast, the testing set test set remained private. As we expect this annotated test dataset to be used for other purposes, the data has been on the Cancer Imaging Archive (TCIA) \citep{tcialnq}.

\subsection{Challenge setup}
The validation phase took place on Grand Challenge, a renowned platform for biomedical challenges, which facilitated automated validation leaderboard management. Participant submissions were automatically assessed using the \texttt{evalutils} and \texttt{MedPy} Python packages. Each participant could make up to three daily submissions on the validation leaderboard. This phase ran from May 1, 2023, to August 30, 2023.

Following best practice guidelines for organizing challenges, the test set remained private to help ensure fairness of the evaluation. Participants were required to containerize their algorithms using Docker, in accordance with Grand Challenge guidelines, and submit their Docker containers for evaluation on the test set. Only one submission was permitted for the test set evaluation. Participants were first encouraged to test their Docker containers on the validation set without submission limits to ensure the algorithms were containerized correctly. If the predictions matched those generated on their machines, participants could submit their algorithms for test set evaluation.

\section{Metrics and evaluation}\label{sec:evaluation}
The choice of metrics to evaluate participants' algorithms and the ranking strategy are crucial for accurate interpretation and reproducibility of results \citep{Maier-Hein2018}. In this section, we adhere to the BIAS best practice recommendations for evaluating challenges \citep{MAIERHEIN2020101796}.

\subsection{Choice of the metrics}
The primary characteristic to optimize for the algorithms is prediction accuracy. Since relying on a single metric for segmentation assessment can result in less robust rankings, we selected two metrics: the Dice similarity coefficient (DSC) and the Average symmetric surface distance (ASSD). These metrics are widely used in previous challenges \citep{CHAOS, antonelli2021medical, dorent2023crossmoda} due to their simplicity, rank stability, and effectiveness in evaluating segmentation accuracy.

Let $S$ represent the predicted binary segmentation mask of the lymph nodes, and  $G$ represent the manual segmentation. The Dice Score coefficient measures the similarity between masks $S$ and $G$ by normalizing the size of their intersection over the average size of the masks:
\begin{equation}
    \DSC(S,G) = \frac{2\sum_{i}S_{i}G_{i}}{\sum_{i}S_{i} + \sum_{i}G_{i}}
\end{equation}
Let $B_{S}$ and $B_{G}$ be the boundaries of the segmentation mask $S$ and the manual segmentation $G$.
The average symmetric surface distance (ASSD)  is calculated as the average of all Euclidean distances (in mm) from points on boundary $B_{S}$ to the boundary $B_{G}$ and vice versa:
\begin{equation}
    \ASSD(S,G) = \frac{\sum_{s_i\in B_{S}}d(s_i, B_{G}) +  \sum_{s_i\in B_{G}}d(s_i, B_{S})}{|B_{S}|+|B_{G}|}
\end{equation}
where $d$ denotes the Euclidean distance.

\subsection{Ranking scheme}
We employed a standard ranking scheme, successfully used in other challenges such as BraTS \citep{BRATS} and crossMoDA~\citep{dorent2023crossmoda}. Teams are ranked for each test case and each metric (DSC and ASSD). In the case of ties, the lowest rank is assigned to the tied values. The overall rank score is calculated by first averaging individual rankings across all cases (cumulative rank) and then averaging these cumulative ranks across all patients for each team. The final team rankings are based on these rank scores. This ranking scheme was defined and published before the challenge began and was available on the Grand Challenge page\footnote{\url{https://lnq2023.grand-challenge.org/}}.

To assess the stability of the ranking scheme, we used the bootstrapping method described by Wiesenfarth et al. (2021). We generated 1,000 bootstrap samples by randomly drawing 100 test cases with replacements from the test set, where each sample retained approximately $63\%$ of distinct cases. The ranking scheme was then applied to each bootstrap sample. We compared the original test set ranking to the rankings from individual bootstrap samples using Kendall's $\tau$, which ranges from $-1$ (reverse order) to $1$ (identical order).

\begin{table*}[ht!]
	\centering
	\caption{Comparisons of the proposed techniques in terms of methodology and implementation details.
	}\label{tab:comparision_teams}
	\resizebox{0.99\textwidth}{!}{
	\begin{tabular}{l c *{5}{c}}
		\toprule
       & \circle{Skeleton Suns-c} \textbf{Skeleton Suns} & \circle{IMR-c} \textbf{IMR} & \circle{CompAI-c} \textbf{CompAI} & \circle{Hilab-c} \textbf{Hilab} & \circle{IMI-c} \textbf{IMI}& \circle{SKJP-c} \textbf{SKJP}\\
		\midrule
    Network architecture & 3D U-Net 
    (nnU-Net) & 3D U-Net 
    (nnU-Net) & 3D U-Net (nnU-Net)
     & 3D U-Net 
    (nnU-Net) &  3D V-Net (nnU-Net)  &2.5D U-Net \\
    \rowcolor{LightGray} 
      &  TCIA CT Lymph Nodes + & TCIA CT Lymph Nodes + & TCIA CT Lymph Nodes + &  & TCIA CT Lymph Nodes + & \\
    \rowcolor{LightGray}
    Use of external& Bouget refinements & Bouget refinements & Bouget refinements & &  Bouget refinements & \\
    \rowcolor{LightGray} 
    datasets & & St. Olavs Hospital  & & & St. Olavs Hospital & \\
    \rowcolor{LightGray}  & &  & NSCLC datasets & \multirow{-4}{*}{$\times$} & & \multirow{-4}{*}{$\times$}  \\
    Cropping & $\times$& Lung and airway & Lung  & Lung & Lung & $\times$\\
    \rowcolor{LightGray} 
    Weakly Supervision & Background  &  & Background   & Partial-supervised & Probabilistic &  \\
    \rowcolor{LightGray}
    Approach & pseudo-labeling & \multirow{-2}{*}{Self-supervision} & pseudo-labeling &  loss functions & atlas & \multirow{-2}{*}{$\times$} \\
     & $\times$ & Small component  & Small component & Small component  &  Small component  & Largest component\\
    \multirow{-2}{*}{Post-processing}& &  removal  &  removal  &  removal &  removal & selection\\ 
    
		\bottomrule
	\end{tabular}
	}
\end{table*}

\section{Participating methods}\label{sec:participating_methods}

A total of 208 teams registered for the challenge, allowing them to download the data. 16 teams from 5 different countries submitted predictions to the validation leaderboard. Among them, 6 teams from 3 countries submitted their containerized algorithm for the evaluation phase. 

In this section, we summarize the methods used by these 6 teams. Each method is assigned a unique color code used in the tables and figures. Brief comparisons of the proposed techniques in terms of methodology and implementation details  are presented in Table~\ref{tab:comparision_teams}. 

\paragraph{\textbf{\circle{Skeleton Suns-c}} Skeleton Suns (1st place, Deissler et al.)}
The authors implemented their method using the nnU-Net architecture \citep{isensee2021nnu} with the extension of a residual encoder \citep{isensee2019attempt}. The overall learning strategy involved a multi-step approach to address the challenges of partially annotated datasets. Initially, the nnU-Net framework was trained using the LNQ challenge dataset \citep{roya_khajavibajestani_2023_7844666}. Due to incomplete annotations in the training data, strategic data enhancement was employed, and an additional fully labeled dataset from TCIA  \citep{roth2015lymphnodedata_seg} with refined segmentations \citep{bouget2023mediastinal} was incorporated to improve training efficacy. To handle incomplete annotations, the team used TotalSegmentator \citep{wasserthal2023totalsegmentator} to identify and label non-lymph node structures, effectively refining the segmentation labels.
The remaining unlabeled regions were excluded from the loss calculation to enable the model to predict unlabeled lymph nodes within these. Additionally, the Bodypartregression toolkit \citep{sarah_schuhegger_2021_5195341} focused the model on the mediastinal region, assigning anatomical scores to each axial slice to exclude non-target areas. These preprocessing steps ensured the model trained effectively on confirmed labels without mislabeling background areas as lymph nodes. The network employed was the '3d fullres' configuration of nnU-Net, with adaptations including adjustments to batch size, patch size, and an extended data augmentation strategy. The training involved a 5-fold cross-validation process, with the final model being an ensemble of these cross-validated models. The ensemble predictions were averaged during inference, though test time augmentations were omitted due to time constraints, slightly degrading performance. Post-processing steps included averaging softmax outputs from the ensemble models to produce the final segmentation masks.

\paragraph{\textbf{\circle{IMR-c}} IMR (2nd place, Zhang et al.)} The team also proposed a semi- and weakly-supervised learning method for automatically segmenting clinically relevant lymph nodes in the mediastinal area of contrast-enhanced CT scans that utilized both partial annotation and full annotation data for two-stage training. First, to better capture the anatomic and semantic representations of mediastinal lymph nodes, a pre-processing approach guided by lung masks and airway maps was used to crop mediastinal VOIs, which included four steps: 1) the lung mask was extracted to obtain the initial VOI boundary of the lung \citep{hofmanninger2020automatic}; 2) this initial VOI was used to segment the airway map \citep{zhang2023deep}; 3) airway voxels were removed in the lung mask to extract the secondary VOI boundary; and 4) the final VOI input was based on the two VOI bounding boxes with margin settings. Then, a two-stage pipeline was designed for semantic segmentation that benefits from partial annotation data. In the first stage, a full-resolution nnU-Net \citep{isensee2021nnu} model was trained initially from scratch with full annotation data from the CT Lymph Node dataset \citep{roth2015lymphnodedata_seg} with refined annotations \citep{bouget2023mediastinal} and the St. Olavs Hospital dataset \citep{bouget2019semantic} based on one-fold of the 5-fold cross-validation for 1000 epochs. In the second stage, the trained model predicted pseudo labels for LNQ training data, which were combined with their partial annotations to produce new lymph node labels. The final model was finetuned using jointly full annotation data and updated partial annotation data for 300 epochs. To exclude non-diseased lymph nodes, each individual component whose volume was less than the volume of a sphere with a radius of 5 mm was removed in the post-processing step.

\paragraph{\textbf{\circle{CompAI-c}} CompAI (3rd place, Fischer et al.)} This team employed a semi- and weakly-supervised approach.  The authors proposed incorporating the TotalSegmentator toolbox \citep{wasserthal2023totalsegmentator} to generate pseudo labels and loss masking for handling incomplete annotations in combination with supervised learning using the nnU-Net framework \citep{isensee2021nnu}. Alongside the challenge training data, the team utilized the fully annotated TCIA CT Lymph Nodes dataset \citep{roth2015lymphnodedata_seg}, providing annotations of all pathologic lymph nodes. Instead of training solely on pathologic lymph nodes, they replaced original annotations with refined annotations containing all visible lymph nodes \citep{bouget2023mediastinal}. Furthermore, to increase the dataset size, the authors included the public lung cancer datasets NSCLC radiomics \citep{aerts2014decoding}, NSCLC radiogenomics \citep{bakr2018radiogenomic}, and NSCLC interobserver \citep{wee2019data}. First, they created ROIs by cropping each volume to the lung bounding box via the TotalSegmentator. To handle incomplete annotations, pseudo labels were generated from TotalSegmentator structures. Those structures should, by definition, not contain any lymph nodes. Labeling those structures as background increased the training supervision significantly. The authors also masked the loss of remaining unlabeled voxels from the training process. Only one nnU-Net instance was trained on the preprocessed data with adjusted hyperparameters for learning rate and intensity clipping. Furthermore, the authors applied a postprocessing step on the segmentation output by discarding all lymph node components with a shortest axis diameter less than 10 mm. In their final challenge report \citep{melba:2024:008:fischer}, the authors show that this postprocessing hurt the overall performance and was based on their misinterpreting the challenge goal.

\paragraph{\textbf{\circle{Hilab-c}} Hilab (4th place, Wang et al.)} This team proposed a framework that combines the techniques of self- and weakly-supervised learning. First, they utilized a self-supervised method named Model Genesis \citep{zhou2021models} to initialize the weights of a VNet-like \citep{milletari2016v} model with one encoder and two decoders. Second, they proposed to add perturbation to the input of one of the decoders to generate variability between the predictions from each decoder. These two predictions were then dynamically mixed to produce better pseudo labels.  To better utilize incomplete annotations, Partial Cross-Entropy (PCE) loss \citep{lee2020scribble2label} and Tversky loss \citep{salehi2017tversky} were applied to this task to balance the supervisory signals of the foreground and background. They also used noise-robust Symmetric Cross-Entropy (SCE) loss \citep{wang2019symmetric} to further extract information from incomplete annotations. To prevent the erroneous foreground voxels in the pseudo labels from misleading the model training, a weighted Cross-Entropy loss was employed following the approach outlined in \cite{zheng2021rectifying}. Only the LNQ challenge training data were used and cropped to the lung region during inference and training. The model was trained in a patch-based manner and inferred using a sliding window strategy. Only the prediction from the decoder without perturbation was used as the inference result. Then, the predicted foreground was refined based on the actual volume of the connected domain, the intensity of the pixel value, and whether it is at the edge of the image. Finally, the prediction was resampled to its original size. The authors later found in their challenge report \citep{melba:2024:017:wang} that although the SCE loss and self-learning method improved the performance on the validation set, degradation on the test set was observed. 

\paragraph{\textbf{\circle{IMI-c}} IMI(5th place, Engelson et al.)} The team proposed an ensemble of five segmentation models based on full-resolution nnU-Net trained on multiple anatomical priors as additional input \citep{isensee2021nnu, SPIE2024}. To address the challenges arising from the weak annotations of the training data, a probabilistic lymph node atlas was registered to the training data to identify regions with high lymph node occurrence, which was then used for loss weighting and post-processing. Further, the authors addressed the heterogeneity of the training data and lymph node appearance by using a strong augmentation called GIN $\&$ IPA introduced by \cite{Ouyang2021}. Preprocessing contained the generation of the anatomical priors using atlas-to-patient registration based on selected segmentation masks from the TotalSegmentator algorithm \citep{wasserthal2023totalsegmentator}, cropping to the lung region, and normalization. For post-processing, the threshold for binarization was lowered according to the lymph node atlas, and segmentation masks below a minimum diameter size of 5 mm were removed. The input data consisted of the weakly-labeled LNQ 2023 training data and publicly available CT Lymph Node dataset \citep{roth2015lymphnodedata_seg} with refined annotations \cite{bouget2023mediastinal} and the St. Olavs Hospital dataset \citep{bouget2019semantic}. The fully annotated data was oversampled during training. After the publication of the fully annotated LNQ 2023 validation and test data, the authors realized that the restriction to a minimum lymph node size degraded segmentation accuracy and that refining on the fully annotated datasets improved performance, as mentioned in their report \citep{melba:2024:009:engelson}.

\begin{table*}[ht!]
	\centering
	\caption{Metrics values and corresponding scores of submission. Median and interquartile values are presented. The best results are given in bold. Arrows indicate the favorable direction of each metric.
	}\label{tab:Scores}
	\resizebox{0.9\textwidth}{!}{
	\begin{tabular}{l c *{4}{c}}
		\toprule
		\multirow{2}{*}{} & \multicolumn{2}{c}{\bf Challenge Rank } & \multicolumn{2}{c}{\bf Lymph Node } \\ 
		
       \cmidrule(lr){2-3} \cmidrule(lr){4-5} 
       & Global Rank $\downarrow$ & Rank Score $\downarrow$ & DSC $(\%)$ $\uparrow$ & ASSD (mm) $\downarrow$ \\
       
		\midrule
    \circle{Skeleton Suns-c} Skeleton Suns & \textbf{1} & \textbf{2.1} & \textbf{71.2 [63.4 - 78.6]} & \textbf{2.95 [2.13 - 5.05]}  \\ 
    \rowcolor{LightGray}
    \circle{IMR-c} IMR & 2 & 2.6 & 70.0 [63.7 - 77.1] & 4.15 [2.92 - 6.3]  \\ 
    \circle{CompAI-c} CompAI & 3 & 2.9 & 69.0 [57.1 - 74.8] & 5.05 [3.12 - 8.79]  \\ 
    \rowcolor{LightGray}
    \circle{Hilab-c} Hilab & 4 & 3.9 & 61.0 [54.0 - 71.1] & 5.92 [3.77 - 8.81]  \\ 
    \circle{IMI-c} IMI& 5 & 3.9 & 61.2 [46.3 - 71.1] & 6.0 [3.84 - 9.43]  \\ 
    \rowcolor{LightGray}
    \circle{SKJP-c} SKJP & 6 & 5.5 & 43.3 [22.6 - 56.6] & 10.4 [6.56 - 15.39]  \\   
		\bottomrule
	\end{tabular}
	}
\end{table*}

\begin{figure*}[ht!]
\centering
\subfloat[Dice Score Similarity (\%)]{\label{fig:lymphdice}{\includegraphics[width=0.48\textwidth]{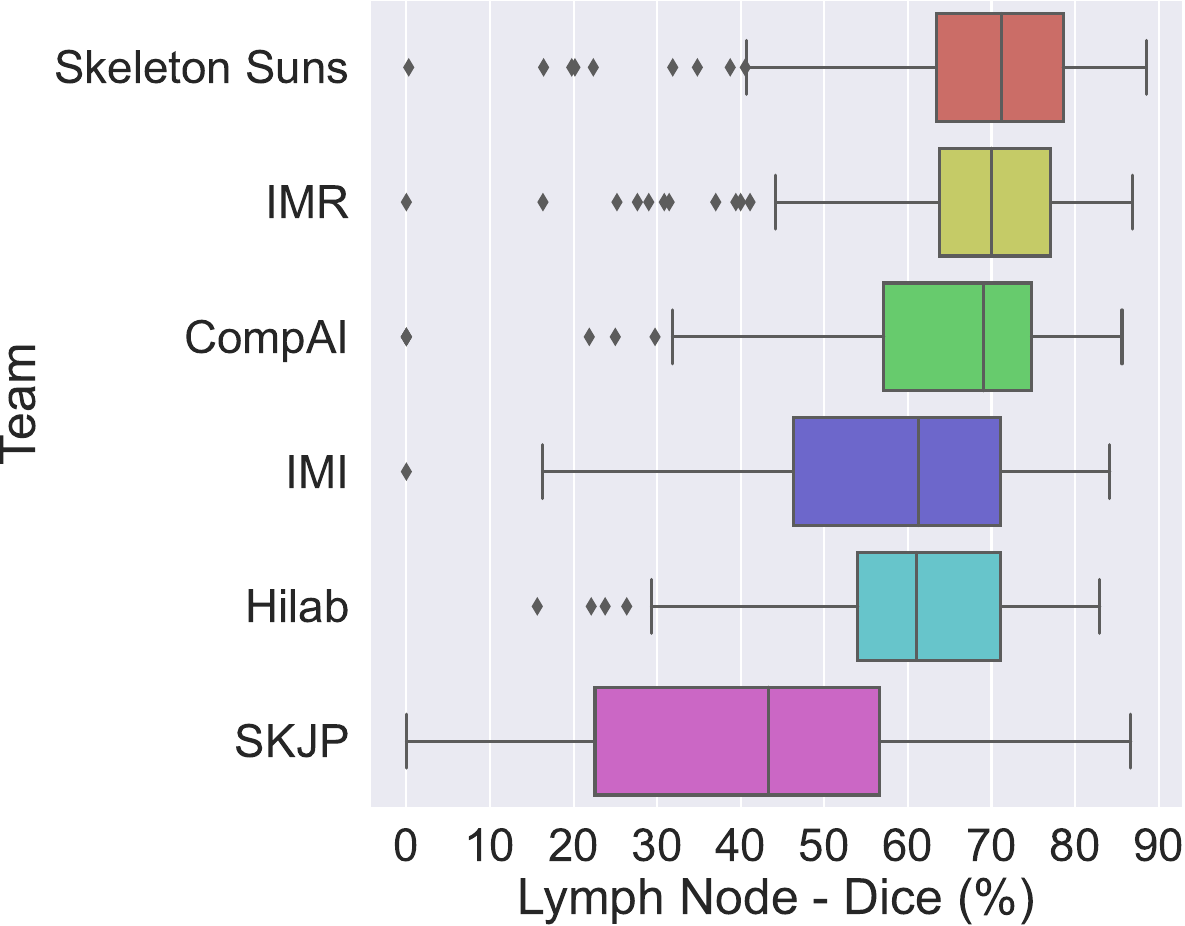}}}\hfill
\subfloat[Average symmetric surface distance (mm)]{\label{fig:lymphassd}{\includegraphics[width=0.48\textwidth]{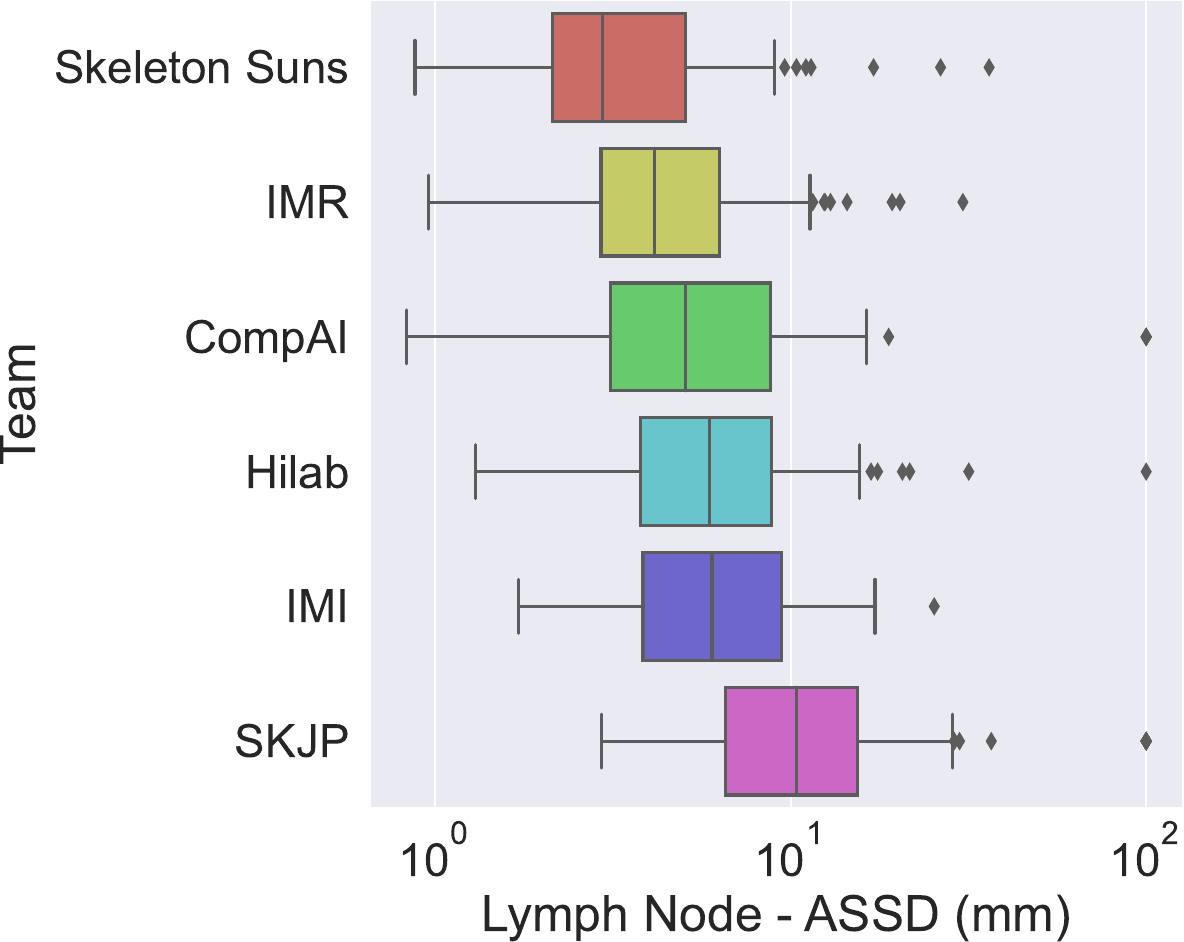}}}
\caption{Box plot of the participants' performance for lymph node segmentation in terms of (a) DSC and (b) ASSD. }
\label{fig:boxlymph}
\end{figure*}

\paragraph{\textbf{\circle{SKJP-c}} SKJP (6th place, Kondo, et al.)} The team used encoder-decoder type deep neural networks. Although the input images are 3D volumes, they use a 2D deep neural network model with multi-slice inputs (2.5D CNNs). Specifically, they used a 2D U-Net as the segmentation network and replaced its encoder part with EfficientNet~\citep{tan2019efficientnet}. In their model, N consecutive slices in each input volume were concatenated and treated as an N channel 2D input image. The input image was then processed to produce one slice in a segmentation mask volume. The input slices were along the transverse plane. During training, slices were randomly selected M times from each volume in the training dataset in each epoch. The loss function was a sum of Dice loss and cross-entropy loss with equal weights. The dataset was split into 353 training samples and 37 validation samples. The optimizer was AdamW and the learning rate was decreased at every epoch with cosine annealing. The model was trained for 100 epochs, and the version with the lowest validation loss was selected as the final model.  Random intensity shifts and random affine transformations were applied as data augmentation. Hyperparameter tuning was conducted, including adjustments to the encoder size, number of slices, and initial learning rate. As a result, EfficientNet-B7 was chosen as the encoder, with 5 slices and an initial learning rate of $10^{-3}$. During inference, each volume was processed slice by slice, and post-processing involved selecting the largest connected component.

\section{Results}\label{sec:results}

Participants were required to submit their algorithm by 20th September 2023. The final results on the testing set were announced during the LNQ workshop at the MICCAI 2023 conference. This section presents the results obtained by the participant teams on the test set and analyses the stability and robustness of the proposed ranking scheme.

\subsection{Overall segmentation performance}
The final scores for the 6 teams are reported in Table~\ref{tab:Scores} in the order in which they ranked. Figures~\ref{fig:lymphdice} and~\ref{fig:lymphassd} show the box plots for each metric (Dice and ASSD) and are color-coded according to the team. 

The winner of the LNQ2023 challenge was the Skeleton Suns, with a rank score of 2.7. Skeleton Suns is the only team with a median ASSD lower than 3 mm. Other teams in the top three also obtained encouraging results with a median DSC greater than $69\%$.  In contrast, the low DSC and high ASSD scores of the team with the lowest rank highlight the complexity of the task.

The top three teams employed a semi- and weakly-supervised approach, combining full supervision using existing fully annotated datasets and weak supervision using the LNQ dataset to train their frameworks. In contrast, methods that only leverage the LNQ weak labels (\circle{Hilab-c}\circle{SKJP-c}) underperformed compared to these semi-supervised approaches. As shown in Table~\ref{tab:Scores}, their medians are significantly lower, and their interquartile ranges (IQRs) are larger. This highlights the effectiveness of leveraging complete annotations in combination with partial annotations to improve model performance.

\begin{figure}[t!]
  \includegraphics[width=0.48\textwidth]{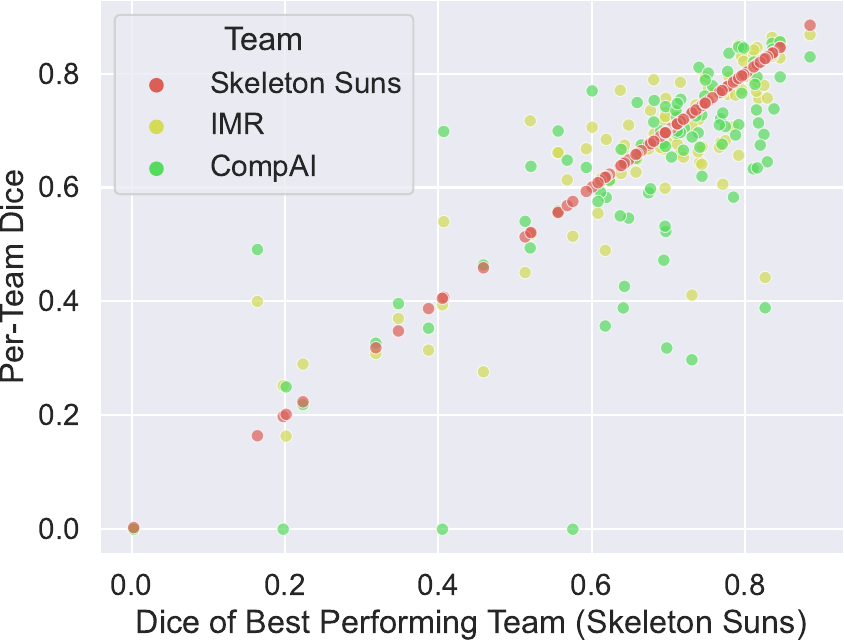}
  \caption{Relationship between the scores for each case between the first, second and third teams. Challenging cases are similar for each team.}
  \label{fig:performance_correlation}
\end{figure} 

\begin{figure*}[ht!]
\centering
\subfloat[Patient sex]{\label{fig:patientsex}{\includegraphics[width=0.4\textwidth]{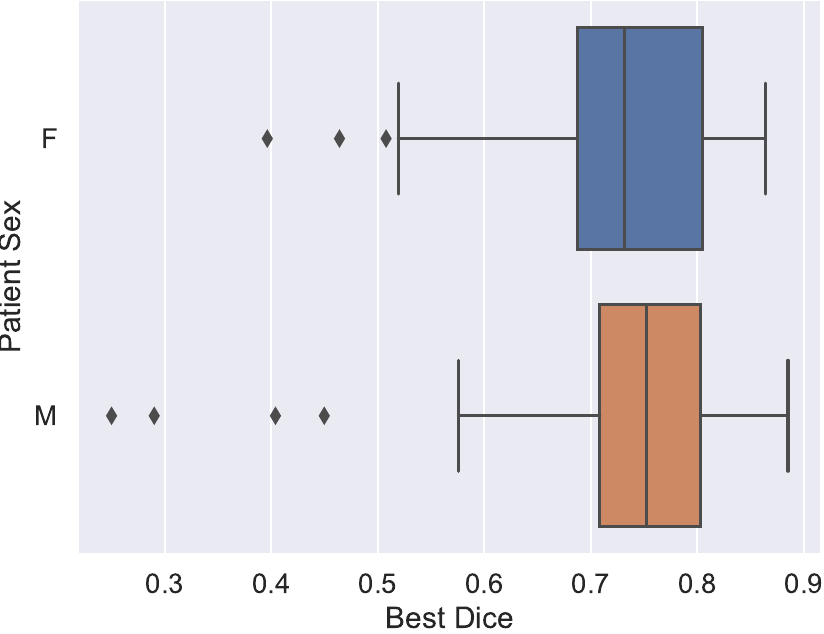}}}\hfill
\subfloat[Patient condition]{\label{fig:patientcondition}{\includegraphics[width=0.57\textwidth]{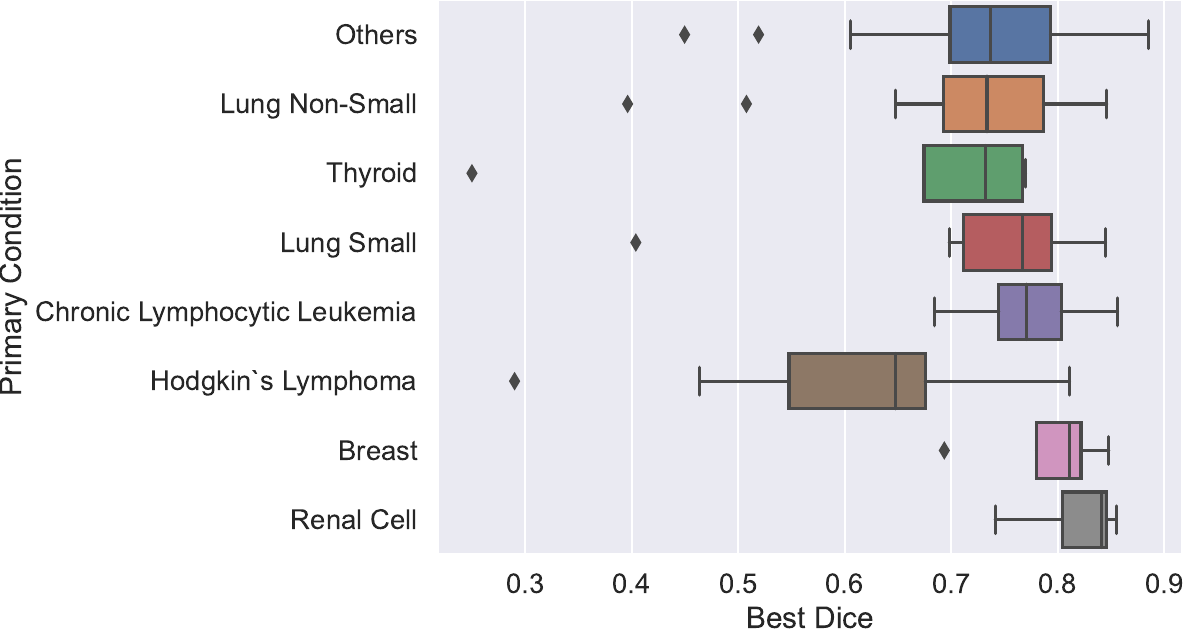}}}
\caption{Box plot of the best performance per (a) patient sex and (b) primary condition. }
\label{fig:boxpatient}
\end{figure*}

\subsection{Analysis of the variations in performance}
While the performance achieved by the top-performing team is high on average, there are patients for whom the best models do not perform well. In this section, we aim to analyze the robustness of the proposed models and identify the more challenging cases.

First, we propose determining whether low-performing cases are similarly distributed across all teams. Specifically, we compare the performance of the second and third teams with that of the top-performing team for each test case. The Figure~\ref{fig:performance_correlation} shows the relationship between the Dice scores of the different techniques, with the x-axis representing the Dice score of the best-performing team and the y-axis representing the Dice scores of the other teams. It is clear that in general challenging cases for the top-performing team are also difficult for the other teams. Additionally, the first two teams exhibit very similar performance for each case, while the third team struggles with some cases successfully segmented by the other two. This suggests that some cases were harder to segment.

To determine if certain patient attributes are predictive of segmentation difficulty, we analyzed the distribution of the best Dice scores across all methods for each case, considering key patient attributes such as cancer type and sex. The results are shown in Figure~\ref{fig:boxpatient}. We found that the best Dice scores had a similar distribution across sexes, as depicted in Figure~\ref{fig:patientsex}. Conversely, Figure~\ref{fig:patientcondition} shows the distribution of scores by patient cancer type. It can be observed that patients with Hodgkin lymphoma had statistically lower scores compared to those with chronic lymphocytic leukemia (CLL) ($p=0.009$), renal cell carcinoma ($p=0.003$), breast cancer ($p=0.008$), and other types ($p=0.27$). This could be attributed to the fact that lymph nodes associated with Hodgkin lymphoma are typically bulky and organized in conglomerates, making their segmentation more challenging.

\subsection{Remarks about the ranking stability}
Several design factors can influence challenge rankings, such as the test set used for validation and the aggregation method applied to these metrics \citep{Maier-Hein2018}. In this section, we analyze and visualize the stability of rankings with respect to these factors.

\begin{figure}[tb!]
  \includegraphics[width=\linewidth]{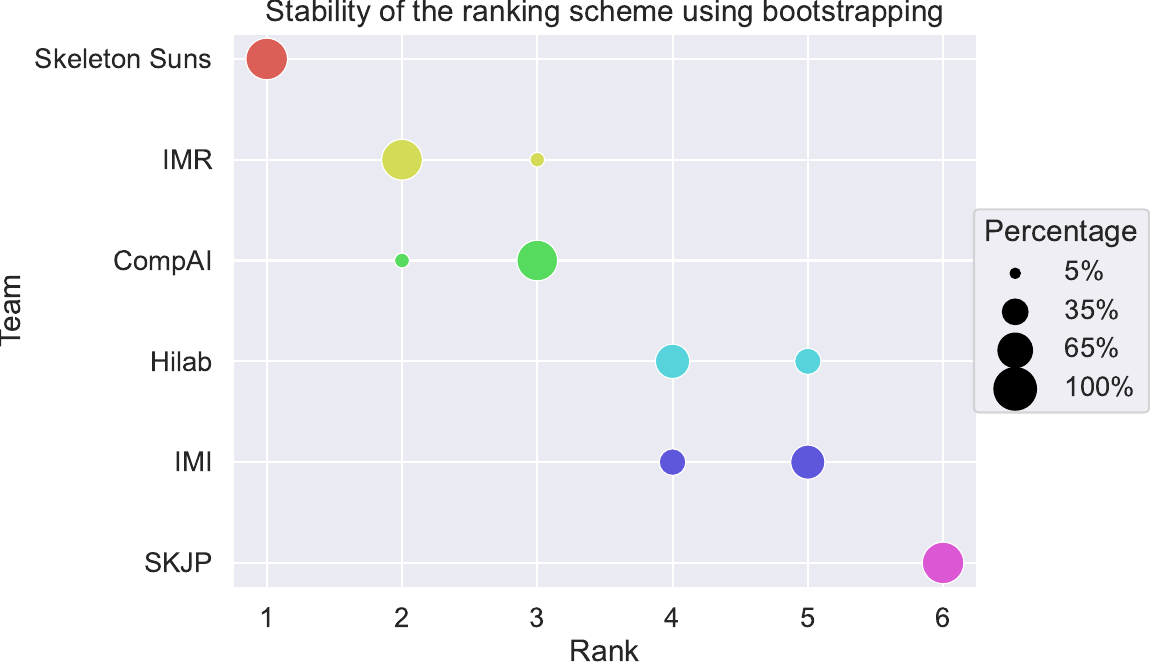}
  \caption{Stability of the proposed ranking scheme for 1000 bootstrap samples.}
  \label{fig:bootstrap_stability}
\end{figure}

To evaluate ranking stability in the context of sampling variability, we adopted the approach described by \cite{Wiesenfarth2021} and used for the crossMoDA challenge~\citep{dorent2023crossmoda}. Following their guidelines, we performed bootstrapping with 1,000 samples to examine the uncertainty and stability of our proposed ranking scheme. The ranking strategy was applied repeatedly to each bootstrap sample. Kendall’s 
$\tau$ was used to quantify the agreement between the original challenge ranking and the rankings derived from the bootstrap samples, yielding values between -1 (indicating reverse ranking order) and 1 (indicating identical ranking order). The median [IQR] Kendall’s  $\tau$ was 1 [$0.87-1$], indicating excellent stability of the ranking scheme. Figure~\ref{fig:bootstrap_stability} presents a blob plot of the bootstrap rankings, confirming the excellent stability, with the winning team consistently ranked first across all bootstrap samples.

We also compared our ranking method with other common aggregation approaches. The main methods are:

\begin{itemize}
    \item Aggregate-then-rank: Metric values are first aggregated (e.g., mean, median) across all test cases for each structure and metric, then ranks are computed for each team. Final ranking scores are derived from aggregating these ranks.
\item Rank-then-aggregate: Ranks are computed for each test case, metric, and structure, then aggregated (e.g., mean, median) to produce a final rank score for each algorithm.
\end{itemize}

\begin{figure}[tb!]
  \includegraphics[width=\linewidth]{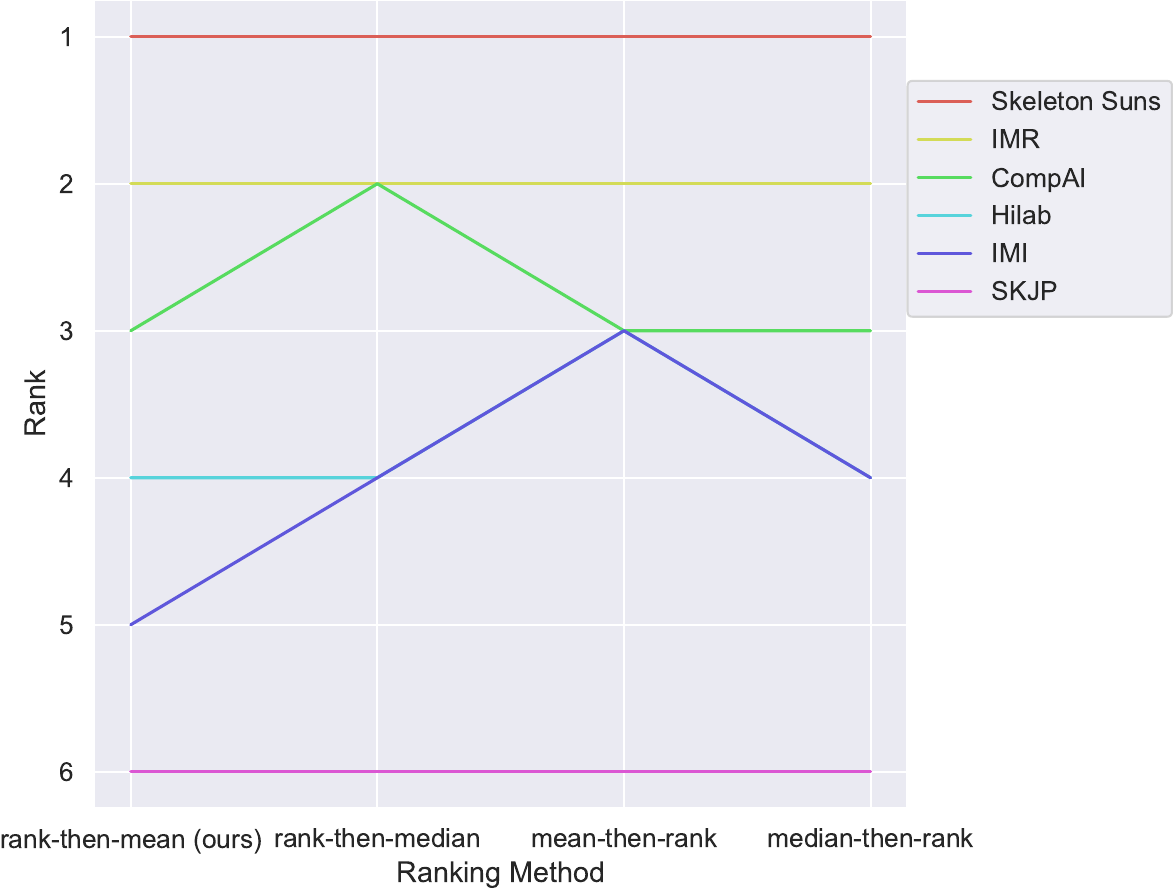}
  \caption{Line plots visualizing rankings robustness across different ranking methods for the brain task. Each algorithm is represented by one colored line. For each ranking method encoded on the x-axis, the height of the line represents the corresponding rank. The lowest rank of tied values is used for ties (equal scores for different teams) }
  \label{fig:ranking_methods}
\end{figure}

Our method uses a rank-then-aggregate approach with the mean as the aggregation technique. We compared this to 1) a rank-then-aggregate approach using the median and 2) aggregate-then-rank approaches using either the mean or the median for metric aggregation. Line plots in Figure~\ref{fig:ranking_methods} illustrate the robustness of rankings across these methods. The ranks remained consistent across all ranking variations, with differences only occurring in cases of ties. Notably, Skeleton Suns was the top-performing team regardless of the ranking approach. This demonstrates that the challenge rankings are stable and can be interpreted with confidence.

\section{Discussion and conclusion}\label{sec:discussion}
In this study, we introduced the LNQ challenge in terms of experimental design, evaluation strategy, proposed methods, and final results. In this section, we discuss the main insights and limitations of the challenge.

\subsection{Full supervision leads to higher performance}
The top-ranked teams all utilized existing fully annotated datasets to train their frameworks. Notably, the top two teams first trained their models without the weakly annotated LNQ data, subsequently refining their models using partial annotations. TotalSegmentator was especially used to remove false positives. In contrast, methods that solely leveraged weak annotations underperformed compared to these fully supervised approaches. This underscores that full supervision still outperforms existing weakly supervised approaches, even when using a smaller amount of training data.

Given that the proposed weakly supervised methods did not close the gap between full and weak supervision, we hope this challenge will continue to serve as a benchmark for developing new weakly supervised approaches.

Additionally, we believe this challenge will contribute to the improvement of existing models trained using full supervision, as the validation and testing datasets currently represent the largest publicly available fully annotated dataset (N=120 scans).

\subsection{Limitations}
This challenge was designed to benchmark new and existing weakly-supervised techniques for lymph node segmentation. In this section, we acknowledge some limitations.

First, there was a noticeable shift in the distribution of cancer types between the training, validation, and testing sets. Despite this  gap, similar performance levels were observed across all cancer types, except for Hodgkin lymphoma. Importantly, the proportion of Hodgkin lymphoma cases was consistent across all three sets, suggesting that the lower performance for this type was due to the inherent difficulty in segmenting its bulky and conglomerated lymph nodes rather than the distribution gap.

The annotation process included selecting lymph nodes considered abnormal, with a short axis length larger than 1 cm. However, it is believed that some lymph nodes around this threshold may have been overlooked. Moreover, no post-processing was performed to remove those with a shorter axis. Some teams attempted to perform this post-processing, which led to degraded results on the test set.

In addition, the weakly annotated training dataset was created by completing volumetric segmentations of lymph nodes that had been selected for measurement during the TIMC clinical trial process, meaning that it may be biased towards the larger or more clinically significant lymph nodes compared to the nodes included in the fully annotated test and validation datasets.

Finally, a single-label map was used for all nodes. We acknowledge that it would have been beneficial to perform instance segmentation, where each node is individually segmented. This represents a potential area for future work.


\subsection{Conclusion}

The LNQ challenge was introduced to propose the first international benchmark for weakly-supervised image segmentation of lymph nodes in 3D CT scans, aimed at advancing the development of weak-supervised segmentation methods in the medical imaging community. By curating a new dataset and providing a standardized evaluation, we facilitated the comparison of different approaches and highlighted the current challenges and limitations of these weakly supervised techniques. Our findings indicate that fully-supervised methods, even when trained on smaller amounts of data, continue to outperform weakly-supervised approaches that leverage larger but partially annotated datasets. This highlights the ongoing need for high-quality, fully annotated data to achieve optimal segmentation performance. Nonetheless, the weakly-supervised methods showed promise, and we believe this challenge will encourage further innovation in this area. Overall, the LNQ challenge provides a valuable resource for the continued development and assessment of lymph node segmentation methods. The fully annotated validation and test sets, in particular, will serve as important assets for future research in lymph node quantification.





\acks{This work was supported by the National Institutes of Health (R01EB032387, P41EB015902, P41EB028741, R01CA235589, 5P30CA006516, 1U24CA258511), National Cancer Data Ecosystem (Task Order No. 413 HHSN26110071 under Contract No. HHSN261201500003l) and EOSS5 and EOSS3 Diversity and Inclusion grants from Chan-Zuckerberg Initiative. R.D. received a Marie
Skłodowska-Curie fellowship No 101154248 (project: SafeREG).  }

%
\ethics{The work follows appropriate ethical standards in conducting research and writing the manuscript, following all applicable laws and regulations regarding treatment of animals or human subjects.
Two IRB protocols were approved at Mass General Brigham covering de-identification of datasets assessed by the Tumor Imaging Metrics Core (TIMC) for the creation of a research data repository for downstream analysis (“Cancer Imaging Research Data Repository for the Lymph Node Quantification Project”, PI: Harris, Protocol Number: 2020P000211) and the annotations made during the clinical trials  (Lymph Node Quantification (LNQ) Project”, PI: Kikinis, Protocol Number: 2020P003754). 
}

\coi{G. Harris and E. Ziegler are co-founders of Yunu, Inc. G. Harris is also at the scientific advisory board of Fovia, Inc. The other authors declare that they have no known competing financial interests or personal relationships that could have appeared to influence the work reported in this paper.}

\data{The data used in the challenge is currently available with registration on the LNQ grand-challenge website at \url{https://lnq2023.grand-challenge.org/} and will be released soon on TCIA as an unrestricted dataset.}

\bibliography{refs.bib}

\begin{thebibliography}{54}
\providecommand{\natexlab}[1]{#1}
\providecommand{\url}[1]{\texttt{#1}}
\expandafter\ifx\csname urlstyle\endcsname\relax
  \providecommand{\doi}[1]{doi: #1}\else
  \providecommand{\doi}{doi: \begingroup \urlstyle{rm}\Url}\fi

\bibitem[Aerts et~al.(2014)Aerts, Velazquez, Leijenaar, Parmar, Grossmann, Carvalho, Bussink, Monshouwer, Haibe-Kains, Rietveld, et~al.]{aerts2014decoding}
H.~Aerts, Emmanuel~Rios Velazquez, Ralph~TH Leijenaar, Chintan Parmar, Patrick Grossmann, Sara Carvalho, Johan Bussink, Ren{\'e} Monshouwer, Benjamin Haibe-Kains, Derek Rietveld, et~al.
\newblock {Decoding tumour phenotype by noninvasive imaging using a quantitative radiomics approach}.
\newblock \emph{{Nature Communications}}, 5\penalty0 (1):\penalty0 4006, 2014.

\bibitem[Antonelli et~al.(2022)Antonelli, Reinke, Bakas, Farahani, Kopp-Schneider, Landman, Litjens, Menze, Ronneberger, Summers, et~al.]{antonelli2021medical}
Michela Antonelli, Annika Reinke, Spyridon Bakas, Keyvan Farahani, Annette Kopp-Schneider, Bennett~A Landman, Geert Litjens, Bjoern Menze, Olaf Ronneberger, Ronald~M Summers, et~al.
\newblock The medical segmentation decathlon.
\newblock \emph{Nature Communications}, 13\penalty0 (1):\penalty0 4128, 2022.

\bibitem[Bakas et~al.(2018)Bakas, Reyes, Jakab, Bauer, Rempfler, Crimi, Shinohara, Berger, Ha, Rozycki, et~al.]{BRATS}
Spyridon Bakas, Mauricio Reyes, Andras Jakab, Stefan Bauer, Markus Rempfler, Alessandro Crimi, Russell~Takeshi Shinohara, Christoph Berger, Sung~Min Ha, Martin Rozycki, et~al.
\newblock {Identifying the best machine learning algorithms for brain tumor segmentation, progression assessment, and overall survival prediction in the BRATS challenge}.
\newblock \emph{arXiv preprint arXiv:1811.02629}, 2018.

\bibitem[Bakr et~al.(2018)Bakr, Gevaert, Echegaray, Ayers, Zhou, Shafiq, Zheng, Benson, Zhang, Leung, et~al.]{bakr2018radiogenomic}
Shaimaa Bakr, Olivier Gevaert, Sebastian Echegaray, Kelsey Ayers, Mu~Zhou, Majid Shafiq, Hong Zheng, Jalen~Anthony Benson, Weiruo Zhang, Ann~NC Leung, et~al.
\newblock {A radiogenomic dataset of non-small cell lung cancer}.
\newblock \emph{{Scientific Data}}, 5\penalty0 (1):\penalty0 1--9, 2018.

\bibitem[Barbu et~al.(2012)Barbu, Suehling, Xu, Liu, Zhou, and Comaniciu]{6033061}
Adrian Barbu, Michael Suehling, Xun Xu, David Liu, S.~Kevin Zhou, and Dorin Comaniciu.
\newblock {Automatic Detection and Segmentation of Lymph Nodes From CT Data}.
\newblock \emph{IEEE Transactions on Medical Imaging}, 31\penalty0 (2):\penalty0 240--250, 2012.

\bibitem[Bouget et~al.(2019)Bouget, J{\o}rgensen, Kiss, Leira, and Lang{\o}]{bouget2019semantic}
David Bouget, Arve J{\o}rgensen, Gabriel Kiss, Haakon~Olav Leira, and Thomas Lang{\o}.
\newblock {Semantic segmentation and detection of mediastinal lymph nodes and anatomical structures in CT data for lung cancer staging}.
\newblock \emph{International Journal of Computer Assisted Radiology and Surgery}, 14:\penalty0 977--986, 2019.

\bibitem[Bouget et~al.(2023)Bouget, Pedersen, Vanel, Leira, and Lang{\o}]{bouget2023mediastinal}
David Bouget, Andr{\'e} Pedersen, Johanna Vanel, Haakon~O Leira, and Thomas Lang{\o}.
\newblock {Mediastinal lymph nodes segmentation using 3D convolutional neural network ensembles and anatomical priors guiding}.
\newblock \emph{Computer Methods in Biomechanics and Biomedical Engineering: Imaging \& Visualization}, 11\penalty0 (1):\penalty0 44--58, 2023.

\bibitem[Cai et~al.(2018)Cai, Tang, Lu, Harrison, Yan, Xiao, Yang, and Summers]{cai2018accurate}
Jinzheng Cai, Youbao Tang, Le~Lu, Adam~P Harrison, Ke~Yan, Jing Xiao, Lin Yang, and Ronald~M Summers.
\newblock {Accurate Weakly-Supervised Deep Lesion Segmentation Using Large-Scale Clinical Annotations: Slice-Propagated 3D Mask Generation from 2D RECIST}.
\newblock In \emph{Medical Image Computing and Computer Assisted Intervention--MICCAI 2018: 21st International Conference, Granada, Spain, September 16-20, 2018, Proceedings, Part IV 11}, pages 396--404. Springer, 2018.

\bibitem[Can et~al.(2018)Can, Chaitanya, Mustafa, Koch, Konukoglu, and Baumgartner]{can2018learning}
Yigit~B Can, Krishna Chaitanya, Basil Mustafa, Lisa~M Koch, Ender Konukoglu, and Christian~F Baumgartner.
\newblock {Learning to segment medical images with scribble-supervision alone}.
\newblock In \emph{Deep Learning in Medical Image Analysis and Multimodal Learning for Clinical Decision Support: 4th International Workshop, DLMIA 2018, and 8th International Workshop, ML-CDS 2018, Held in Conjunction with MICCAI 2018, Granada, Spain, September 20, 2018, Proceedings 4}, pages 236--244. Springer, 2018.

\bibitem[Cardoso et~al.(2022)Cardoso, Li, Brown, Ma, Kerfoot, Wang, Murrey, Myronenko, Zhao, Yang, Nath, He, Xu, Hatamizadeh, Myronenko, Zhu, Liu, Zheng, Tang, Yang, Zephyr, Hashemian, Alle, Darestani, Budd, Modat, Vercauteren, Wang, Li, Hu, Fu, Gorman, Johnson, Genereaux, Erdal, Gupta, Diaz-Pinto, Dourson, Maier-Hein, Jaeger, Baumgartner, Kalpathy-Cramer, Flores, Kirby, Cooper, Roth, Xu, Bericat, Floca, Zhou, Shuaib, Farahani, Maier-Hein, Aylward, Dogra, Ourselin, and Feng]{cardoso2022monai}
M.~Jorge Cardoso, Wenqi Li, Richard Brown, Nic Ma, Eric Kerfoot, Yiheng Wang, Benjamin Murrey, Andriy Myronenko, Can Zhao, Dong Yang, Vishwesh Nath, Yufan He, Ziyue Xu, Ali Hatamizadeh, Andriy Myronenko, Wentao Zhu, Yun Liu, Mingxin Zheng, Yucheng Tang, Isaac Yang, Michael Zephyr, Behrooz Hashemian, Sachidanand Alle, Mohammad~Zalbagi Darestani, Charlie Budd, Marc Modat, Tom Vercauteren, Guotai Wang, Yiwen Li, Yipeng Hu, Yunguan Fu, Benjamin Gorman, Hans Johnson, Brad Genereaux, Barbaros~S. Erdal, Vikash Gupta, Andres Diaz-Pinto, Andre Dourson, Lena Maier-Hein, Paul~F. Jaeger, Michael Baumgartner, Jayashree Kalpathy-Cramer, Mona Flores, Justin Kirby, Lee A.~D. Cooper, Holger~R. Roth, Daguang Xu, David Bericat, Ralf Floca, S.~Kevin Zhou, Haris Shuaib, Keyvan Farahani, Klaus~H. Maier-Hein, Stephen Aylward, Prerna Dogra, Sebastien Ourselin, and Andrew Feng.
\newblock {MONAI: An open-source framework for deep learning in healthcare}, 2022.

\bibitem[Dorent et~al.(2020)Dorent, Joutard, Shapey, Bisdas, Kitchen, Bradford, Saeed, Modat, Ourselin, and Vercauteren]{dorent2020scribble}
Reuben Dorent, Samuel Joutard, Jonathan Shapey, Sotirios Bisdas, Neil Kitchen, Robert Bradford, Shakeel Saeed, Marc Modat, S{\'e}bastien Ourselin, and Tom Vercauteren.
\newblock Scribble-based domain adaptation via co-segmentation.
\newblock In \emph{Medical Image Computing and Computer Assisted Intervention--MICCAI 2020: 23rd International Conference, Lima, Peru, October 4--8, 2020, Proceedings, Part I 23}, pages 479--489. Springer, 2020.

\bibitem[Dorent et~al.(2021{\natexlab{a}})Dorent, Booth, Li, Sudre, Kafiabadi, Cardoso, Ourselin, and Vercauteren]{dorent2021learning}
Reuben Dorent, Thomas Booth, Wenqi Li, Carole~H Sudre, Sina Kafiabadi, Jorge Cardoso, Sebastien Ourselin, and Tom Vercauteren.
\newblock Learning joint segmentation of tissues and brain lesions from task-specific hetero-modal domain-shifted datasets.
\newblock \emph{Medical Image Analysis}, 67:\penalty0 101862, 2021{\natexlab{a}}.

\bibitem[Dorent et~al.(2021{\natexlab{b}})Dorent, Joutard, Shapey, Kujawa, Modat, Ourselin, and Vercauteren]{dorent2021inter}
Reuben Dorent, Samuel Joutard, Jonathan Shapey, Aaron Kujawa, Marc Modat, S{\'e}bastien Ourselin, and Tom Vercauteren.
\newblock {Inter extreme points geodesics for end-to-end weakly supervised image segmentation}.
\newblock In \emph{Medical Image Computing and Computer Assisted Intervention--MICCAI 2021: 24th International Conference, Strasbourg, France, September 27--October 1, 2021, Proceedings, Part II 24}, pages 615--624. Springer, 2021{\natexlab{b}}.

\bibitem[Dorent et~al.(2023)Dorent, Kujawa, Ivory, Bakas, Rieke, Joutard, Glocker, Cardoso, Modat, Batmanghelich, et~al.]{dorent2023crossmoda}
Reuben Dorent, Aaron Kujawa, Marina Ivory, Spyridon Bakas, Nicola Rieke, Samuel Joutard, Ben Glocker, Jorge Cardoso, Marc Modat, Kayhan Batmanghelich, et~al.
\newblock {CrossMoDA 2021 challenge: Benchmark of cross-modality domain adaptation techniques for vestibular schwannoma and cochlea segmentation}.
\newblock \emph{Medical Image Analysis}, 83:\penalty0 102628, 2023.

\bibitem[Eisenhauer et~al.(2009)Eisenhauer, Therasse, Bogaerts, Schwartz, Sargent, Ford, Dancey, Arbuck, Gwyther, Mooney, et~al.]{eisenhauer2009new}
Elizabeth~A Eisenhauer, Patrick Therasse, Jan Bogaerts, Lawrence~H Schwartz, Danielle Sargent, Robert Ford, Janet Dancey, S~Arbuck, Steve Gwyther, Margaret Mooney, et~al.
\newblock {New response evaluation criteria in solid tumours: revised RECIST guideline (version 1.1)}.
\newblock \emph{European Journal of Cancer}, 45\penalty0 (2):\penalty0 228--247, 2009.

\bibitem[Engelson et~al.(2024{\natexlab{a}})Engelson, Ehrhardt, Kepp, Niemeijer, and Handels]{melba:2024:009:engelson}
Sofija Engelson, Jan Ehrhardt, Timo Kepp, Joshua Niemeijer, and Heinz Handels.
\newblock Lnq challenge 2023: Learning mediastinal lymph node segmentation with a probabilistic lymph node atlas.
\newblock \emph{Machine Learning for Biomedical Imaging}, 2:\penalty0 817--833, 2024{\natexlab{a}}.
\newblock ISSN 2766-905X.
\newblock \doi{https://doi.org/10.59275/j.melba.2024-f95c}.
\newblock URL \url{https://melba-journal.org/2024:009}.

\bibitem[Engelson et~al.(2024{\natexlab{b}})Engelson, Ehrhardt, Niemeijer, Schierholz, Berkel, Elser, and Handels]{SPIE2024}
Sofija Engelson, Jan Ehrhardt, Joshua Niemeijer, Stefanie Schierholz, Lennart Berkel, Malte~Maria Elser, Yannic~Sieren, and Heinz Handels.
\newblock {Comparison of Anatomical Priors for Learning-based Neural Network Guidance for Mediastinal Lymph Node Segmentation}.
\newblock In Weijie Chen and Susan~M. Astley, editors, \emph{Medical Imaging 2024: Computer-Aided Diagnosis}, volume 12927, page 1292719. International Society for Optics and Photonics, SPIE, 2024{\natexlab{b}}.

\bibitem[Fedorov et~al.(2012)Fedorov, Beichel, Kalpathy-Cramer, Finet, Fillion-Robin, Pujol, Bauer, Jennings, Fennessy, Sonka, et~al.]{fedorov20123d}
Andriy Fedorov, Reinhard Beichel, Jayashree Kalpathy-Cramer, Julien Finet, Jean-Christophe Fillion-Robin, Sonia Pujol, Christian Bauer, Dominique Jennings, Fiona Fennessy, Milan Sonka, et~al.
\newblock {3D Slicer as an image computing platform for the Quantitative Imaging Network}.
\newblock \emph{Magnetic resonance imaging}, 30\penalty0 (9):\penalty0 1323--1341, 2012.

\bibitem[Feulner et~al.(2013)Feulner, Zhou, Hammon, Hornegger, and Comaniciu]{feulner2013lymph}
Johannes Feulner, S~Kevin Zhou, Matthias Hammon, Joachim Hornegger, and Dorin Comaniciu.
\newblock {Lymph node detection and segmentation in chest CT data using discriminative learning and a spatial prior}.
\newblock \emph{Medical Image Analysis}, 17\penalty0 (2):\penalty0 254--270, 2013.

\bibitem[Fischer et~al.(2024)Fischer, Kiechle, Lang, Peeken, and Schnabel]{melba:2024:008:fischer}
Stefan~M. Fischer, Johannes Kiechle, Daniel~M. Lang, Jan~C. Peeken, and Julia~A. Schnabel.
\newblock Mask the unknown: Assessing different strategies to handle weak annotations in the miccai2023 mediastinal lymph node quantification challenge.
\newblock \emph{Machine Learning for Biomedical Imaging}, 2:\penalty0 798--816, 2024.
\newblock ISSN 2766-905X.

\bibitem[Hofmanninger et~al.(2020)Hofmanninger, Prayer, Pan, R{\"o}hrich, Prosch, and Langs]{hofmanninger2020automatic}
Johannes Hofmanninger, Forian Prayer, Jeanny Pan, Sebastian R{\"o}hrich, Helmut Prosch, and Georg Langs.
\newblock {Automatic lung segmentation in routine imaging is primarily a data diversity problem, not a methodology problem}.
\newblock \emph{European Radiology Experimental}, 4:\penalty0 1--13, 2020.

\bibitem[Idris et~al.(2024)Idris, Somarouthu, Jacene, LaCasce, Ziegler, Pieper, Khajavi, Dorent, Pujol, Kikinis, and Harris]{tcialnq}
Tagwa Idris, Bhanusupriya Somarouthu, Heather Jacene, Ann LaCasce, Erik Ziegler, Steve Pieper, Roya Khajavi, Reuben Dorent, Sonia Pujol, Ron Kikinis, and Gordon Harris.
\newblock {Mediastinal Lymph Node Quantification (LNQ): Segmentation of Heterogeneous CT Data}.
\newblock \emph{The Cancer Imaging Archive}, 2024.

\bibitem[Isensee and Maier-Hein(2019)]{isensee2019attempt}
Fabian Isensee and Klaus~H Maier-Hein.
\newblock {An attempt at beating the 3D U-Net}.
\newblock \emph{arXiv preprint arXiv:1908.02182}, 2019.

\bibitem[Isensee et~al.(2021)Isensee, Jaeger, Kohl, Petersen, and Maier-Hein]{isensee2021nnu}
Fabian Isensee, Paul~F Jaeger, Simon~AA Kohl, Jens Petersen, and Klaus~H Maier-Hein.
\newblock {nnU-Net: a self-configuring method for deep learning-based biomedical image segmentation}.
\newblock \emph{Nature methods}, 18\penalty0 (2):\penalty0 203--211, 2021.

\bibitem[Kavur et~al.(2021)Kavur, Gezer, Barış, Aslan, Conze, Groza, Pham, Chatterjee, Ernst, Özkan, Baydar, Lachinov, Han, Pauli, Isensee, Perkonigg, Sathish, Rajan, Sheet, Dovletov, Speck, Nürnberger, Maier-Hein, {Bozdağı Akar}, Ünal, Dicle, and Selver]{CHAOS}
A.~Emre Kavur, N.~Sinem Gezer, Mustafa Barış, Sinem Aslan, Pierre-Henri Conze, Vladimir Groza, Duc~Duy Pham, Soumick Chatterjee, Philipp Ernst, Savaş Özkan, Bora Baydar, Dmitry Lachinov, Shuo Han, Josef Pauli, Fabian Isensee, Matthias Perkonigg, Rachana Sathish, Ronnie Rajan, Debdoot Sheet, Gurbandurdy Dovletov, Oliver Speck, Andreas Nürnberger, Klaus~H. Maier-Hein, Gözde {Bozdağı Akar}, Gözde Ünal, Oğuz Dicle, and M.~Alper Selver.
\newblock {CHAOS Challenge - combined (CT-MR) healthy abdominal organ segmentation}.
\newblock \emph{Medical Image Analysis}, 69:\penalty0 101950, 2021.
\newblock ISSN 1361-8415.

\bibitem[Kervadec et~al.(2020)Kervadec, Dolz, Wang, Granger, and {Ben Ayed}]{pmlr-v121-kervadec20a}
Hoel Kervadec, Jose Dolz, Shanshan Wang, Eric Granger, and Ismail {Ben Ayed}.
\newblock {Bounding boxes for weakly supervised segmentation: Global constraints get close to full supervision}.
\newblock In Tal Arbel, Ismail Ben~Ayed, Marleen de~Bruijne, Maxime Descoteaux, Herve Lombaert, and Christopher Pal, editors, \emph{Proceedings of the Third Conference on Medical Imaging with Deep Learning}, volume 121 of \emph{Proceedings of Machine Learning Research}, pages 365--381. PMLR, 06--08 Jul 2020.

\bibitem[Khajavi et~al.(2023)Khajavi, Pieper, Ziegler, Idris, Dorent, Somarouthu, Pujol, LaCasce, Jacene, Harris, and Kikinis]{roya_khajavibajestani_2023_7844666}
Roya Khajavi, Steve Pieper, Erik Ziegler, Tagwa Idris, Reuben Dorent, Bhanusupriya Somarouthu, Sonia Pujol, Ann LaCasce, Heather Jacene, Gordon Harris, and Ron Kikinis.
\newblock {Mediastinal Lymph Node Quantification (LNQ): Segmentation of Heterogeneous CT Data}.
\newblock \emph{Zenodo}, April 2023.
\newblock \doi{10.5281/zenodo.7844666}.

\bibitem[Lee and Jeong(2020)]{lee2020scribble2label}
Hyeonsoo Lee and Won-Ki Jeong.
\newblock {Scribble2label: Scribble-supervised cell segmentation via self-generating pseudo-labels with consistency}.
\newblock In \emph{Medical Image Computing and Computer Assisted Intervention--MICCAI 2020: 23rd International Conference, Lima, Peru, October 4--8, 2020, Proceedings, Part I 23}, pages 14--23. Springer, 2020.

\bibitem[Li and Xia(2020)]{li2020deep}
Zhe Li and Yong Xia.
\newblock {Deep reinforcement learning for weakly-supervised lymph node segmentation in CT images}.
\newblock \emph{IEEE Journal of Biomedical and Health Informatics}, 25\penalty0 (3):\penalty0 774--783, 2020.

\bibitem[Maier-Hein et~al.(2018)Maier-Hein, Eisenmann, Reinke, Onogur, Stankovic, Scholz, Arbel, Bogunovic, Bradley, Carass, Feldmann, Frangi, Full, van Ginneken, Hanbury, Honauer, Kozubek, Landman, M{\"a}rz, Maier, Maier-Hein, Menze, M{\"u}ller, Neher, Niessen, Rajpoot, Sharp, Sirinukunwattana, Speidel, Stock, Stoyanov, Taha, van~der Sommen, Wang, Weber, Zheng, Jannin, and Kopp-Schneider]{Maier-Hein2018}
Lena Maier-Hein, Matthias Eisenmann, Annika Reinke, Sinan Onogur, Marko Stankovic, Patrick Scholz, Tal Arbel, Hrvoje Bogunovic, Andrew~P. Bradley, Aaron Carass, Carolin Feldmann, Alejandro~F. Frangi, Peter~M. Full, Bram van Ginneken, Allan Hanbury, Katrin Honauer, Michal Kozubek, Bennett~A. Landman, Keno M{\"a}rz, Oskar Maier, Klaus Maier-Hein, Bjoern~H. Menze, Henning M{\"u}ller, Peter~F. Neher, Wiro Niessen, Nasir Rajpoot, Gregory~C. Sharp, Korsuk Sirinukunwattana, Stefanie Speidel, Christian Stock, Danail Stoyanov, Abdel~Aziz Taha, Fons van~der Sommen, Ching-Wei Wang, Marc-Andr{\'e} Weber, Guoyan Zheng, Pierre Jannin, and Annette Kopp-Schneider.
\newblock {Why rankings of biomedical image analysis competitions should be interpreted with care}.
\newblock \emph{{Nature Communications}}, 9\penalty0 (1):\penalty0 5217, Dec 2018.
\newblock ISSN 2041-1723.

\bibitem[Maier-Hein et~al.(2020)Maier-Hein, Reinke, Kozubek, Martel, Arbel, Eisenmann, Hanbury, Jannin, Müller, Onogur, Saez-Rodriguez, {van Ginneken}, Kopp-Schneider, and Landman]{MAIERHEIN2020101796}
Lena Maier-Hein, Annika Reinke, Michal Kozubek, Anne~L. Martel, Tal Arbel, Matthias Eisenmann, Allan Hanbury, Pierre Jannin, Henning Müller, Sinan Onogur, Julio Saez-Rodriguez, Bram {van Ginneken}, Annette Kopp-Schneider, and Bennett~A. Landman.
\newblock {BIAS: Transparent reporting of biomedical image analysis challenges}.
\newblock \emph{Medical Image Analysis}, 66:\penalty0 101796, 2020.
\newblock ISSN 1361-8415.

\bibitem[Mehrtash et~al.(2024)Mehrtash, Ziegler, Idris, Somarouthu, Urban, LaCasce, Jacene, Van Den~Abbeele, Pieper, Harris, et~al.]{mehrtash2024evaluation}
Alireza Mehrtash, Erik Ziegler, Tagwa Idris, Bhanusupriya Somarouthu, Trinity Urban, Ann~S LaCasce, Heather Jacene, Annick~D Van Den~Abbeele, Steve Pieper, Gordon Harris, et~al.
\newblock {Evaluation of mediastinal lymph node segmentation of heterogeneous CT data with full and weak supervision}.
\newblock \emph{Computerized Medical Imaging and Graphics}, 111:\penalty0 102312, 2024.

\bibitem[Milletari et~al.(2016)Milletari, Navab, and Ahmadi]{milletari2016v}
Fausto Milletari, Nassir Navab, and Seyed-Ahmad Ahmadi.
\newblock {V-net: Fully convolutional neural networks for volumetric medical image segmentation}.
\newblock In \emph{2016 fourth international conference on 3D vision (3DV)}, pages 565--571. Ieee, 2016.

\bibitem[Ouyang et~al.(2021)Ouyang, Chen, Li, Li, Qin, Bai, and Rueckert]{Ouyang2021}
Cheng Ouyang, Chen Chen, Surui Li, Zeju Li, Chen Qin, Wenjia Bai, and Daniel Rueckert.
\newblock {Causality-Inspired Single-Source Domain Generalization for Medical Image Segmentation}.
\newblock \emph{IEEE Transactions on Medical Imaging}, 42:\penalty0 1095--1106, 2021.

\bibitem[Ouyang et~al.(2019)Ouyang, Xue, Zhan, Zhou, Wang, Zhou, Wang, and Cheng]{ouyang2019weakly}
Xi~Ouyang, Zhong Xue, Yiqiang Zhan, Xiang~Sean Zhou, Qingfeng Wang, Ying Zhou, Qian Wang, and Jie-Zhi Cheng.
\newblock {Weakly supervised segmentation framework with uncertainty: A study on pneumothorax segmentation in chest X-ray}.
\newblock In \emph{Medical Image Computing and Computer Assisted Intervention--MICCAI 2019: 22nd International Conference, Shenzhen, China, October 13--17, 2019, Proceedings, Part VI 22}, pages 613--621. Springer, 2019.

\bibitem[Pieper et~al.(2004)Pieper, Halle, and Kikinis]{pieper20043d}
Steve Pieper, Michael Halle, and Ron Kikinis.
\newblock {3D Slicer}.
\newblock In \emph{2004 2nd IEEE international symposium on biomedical imaging: nano to macro (IEEE Cat No. 04EX821)}, pages 632--635. IEEE, 2004.

\bibitem[Roth et~al.(2015)Roth, Lu, Seff, Cherry, Hoffman, Wang, Liu, Turkbey, and Summers]{roth2015lymphnodedata_seg}
Holger Roth, Le~Lu, Ari Seff, Kevin~M Cherry, Joanne Hoffman, Shijun Wang, Jiamin Liu, Evrim Turkbey, and Ronald~M. Summers.
\newblock {A new 2.5 D representation for lymph node detection in CT (CT Lymph Nodes)}.
\newblock \emph{The Cancer Imaging Archive}, 2015.

\bibitem[{Roth, Holger and Zhang, Ling and Yang, Dong and Milletari, Fausto and Xu, Ziyue and Wang, Xiaosong and Xu, Daguang}(2019)]{roth2019weakly}
{Roth, Holger and Zhang, Ling and Yang, Dong and Milletari, Fausto and Xu, Ziyue and Wang, Xiaosong and Xu, Daguang}.
\newblock {Weakly supervised segmentation from extreme points}.
\newblock In \emph{Large-Scale Annotation of Biomedical Data and Expert Label Synthesis and Hardware Aware Learning for Medical Imaging and Computer Assisted Intervention: International Workshops, LABELS 2019, HAL-MICCAI 2019, and CuRIOUS 2019, Held in Conjunction with MICCAI 2019, Shenzhen, China, October 13 and 17, 2019, Proceedings 4}, pages 42--50. Springer, 2019.

\bibitem[Salehi et~al.(2017)Salehi, Erdogmus, and Gholipour]{salehi2017tversky}
Seyed Sadegh~Mohseni Salehi, Deniz Erdogmus, and Ali Gholipour.
\newblock {Tversky loss function for image segmentation using 3D fully convolutional deep networks}.
\newblock In \emph{International workshop on machine learning in medical imaging}, pages 379--387. Springer, 2017.

\bibitem[Schuhegger(2021)]{sarah_schuhegger_2021_5195341}
Sarah Schuhegger.
\newblock {MIC-DKFZ BodyPartRegression}.
\newblock \emph{Zenodo}, August 2021.
\newblock \doi{10.5281/zenodo.5195341}.

\bibitem[Stapleford et~al.(2010)Stapleford, Lawson, Perkins, Edelman, Davis, McDonald, Waller, Schreibmann, and Fox]{STAPLEFORD2010959}
Liza~J. Stapleford, Joshua~D. Lawson, Charles Perkins, Scott Edelman, Lawrence Davis, Mark~W. McDonald, Anthony Waller, Eduard Schreibmann, and Tim Fox.
\newblock {Evaluation of Automatic Atlas-Based Lymph Node Segmentation for Head-and-Neck Cancer}.
\newblock \emph{International Journal of Radiation Oncology - Biology - Physics}, 77\penalty0 (3):\penalty0 959--966, 2010.
\newblock ISSN 0360-3016.

\bibitem[Tan and Le(2019)]{tan2019efficientnet}
Mingxing Tan and Quoc Le.
\newblock {Efficientnet: Rethinking model scaling for convolutional neural networks}.
\newblock In \emph{International Conference on Machine Learning}, pages 6105--6114. PMLR, 2019.

\bibitem[Tan et~al.(2018)Tan, Lu, Bonde, Wang, Qi, Schwartz, and Zhao]{tan2018}
Yongqiang Tan, Lin Lu, Apurva Bonde, Deling Wang, Jing Qi, Lawrence~H. Schwartz, and Binsheng Zhao.
\newblock {Lymph node segmentation by dynamic programming and active contours}.
\newblock \emph{Medical Physics}, 45\penalty0 (5):\penalty0 2054--2062, 2018.

\bibitem[Wang et~al.(2024)Wang, Qu, Luo, Liao, Zhang, and Wang]{melba:2024:017:wang}
Litingyu Wang, Yijie Qu, Xiangde Luo, Wenjun Liao, Shichuan Zhang, and Guotai Wang.
\newblock Weakly supervised lymph nodes segmentation based on partial instance annotations with pre-trained dual-branch network and pseudo label learning.
\newblock \emph{Machine Learning for Biomedical Imaging}, 2:\penalty0 1030--1047, 2024.
\newblock ISSN 2766-905X.
\newblock \doi{https://doi.org/10.59275/j.melba.2024-489g}.
\newblock URL \url{https://melba-journal.org/2024:017}.

\bibitem[Wang et~al.(2019)Wang, Ma, Chen, Luo, Yi, and Bailey]{wang2019symmetric}
Yisen Wang, Xingjun Ma, Zaiyi Chen, Yuan Luo, Jinfeng Yi, and James Bailey.
\newblock {Symmetric cross entropy for robust learning with noisy labels}.
\newblock In \emph{IEEE/CVF International Conference on Computer Vision}, pages 322--330, 2019.

\bibitem[Wasserthal et~al.(2023)Wasserthal, Breit, Meyer, Pradella, Hinck, Sauter, Heye, Boll, Cyriac, Yang, Bach, and Segeroth]{wasserthal2023totalsegmentator}
Jakob Wasserthal, Hanns-Christian Breit, Manfred~T. Meyer, Maurice Pradella, Daniel Hinck, Alexander~W. Sauter, Tobias Heye, Daniel~T. Boll, Joshy Cyriac, Shan Yang, Michael Bach, and Martin Segeroth.
\newblock {TotalSegmentator: Robust Segmentation of 104 Anatomic Structures in CT Images}.
\newblock \emph{Radiology: Artificial Intelligence}, 5\penalty0 (5):\penalty0 e230024, 2023.

\bibitem[Wee et~al.(2019)Wee, Aerts, Kalendralis, and Dekker]{wee2019data}
L~Wee, HJL Aerts, P~Kalendralis, and A~Dekker.
\newblock {Data From NSCLC-Radiomics-Interobserver1 [Data set]}.
\newblock \emph{The Cancer Imaging Archive}, 10, 2019.

\bibitem[Wiesenfarth et~al.(2021)Wiesenfarth, Reinke, Landman, Eisenmann, Saiz, Cardoso, Maier-Hein, and Kopp-Schneider]{Wiesenfarth2021}
Manuel Wiesenfarth, Annika Reinke, Bennett~A. Landman, Matthias Eisenmann, Laura~Aguilera Saiz, M.~Jorge Cardoso, Lena Maier-Hein, and Annette Kopp-Schneider.
\newblock {Methods and open-source toolkit for analyzing and visualizing challenge results}.
\newblock \emph{Scientific Reports}, 11\penalty0 (1):\penalty0 2369, Jan 2021.
\newblock ISSN 2045-2322.

\bibitem[Zhang et~al.(2023)Zhang, Zhang, Gu, and Yang]{zhang2023deep}
Hanxiao Zhang, Minghui Zhang, Yun Gu, and Guang-Zhong Yang.
\newblock {Deep anatomy learning for lung airway and artery-vein modeling with contrast-enhanced CT synthesis}.
\newblock \emph{International Journal of Computer Assisted Radiology and Surgery}, 18\penalty0 (7):\penalty0 1287--1294, 2023.

\bibitem[Zhang and Zhuang(2022)]{zhang2022cyclemix}
Ke~Zhang and Xiahai Zhuang.
\newblock {Cyclemix: A holistic strategy for medical image segmentation from scribble supervision}.
\newblock In \emph{Proceedings of the IEEE/CVF Conference on Computer Vision and Pattern Recognition}, pages 11656--11665, 2022.

\bibitem[Zhang et~al.(2020)Zhang, Zhou, Zhao, Man, Liu, and Yao]{zhang2020survey}
Man Zhang, Yong Zhou, Jiaqi Zhao, Yiyun Man, Bing Liu, and Rui Yao.
\newblock A survey of semi-and weakly supervised semantic segmentation of images.
\newblock \emph{{Artificial Intelligence Review}}, 53:\penalty0 4259--4288, 2020.

\bibitem[Zhao et~al.(2009)Zhao, James, Moskowitz, Guo, Ginsberg, Lefkowitz, Qin, Riely, Kris, and Schwartz]{zhao2009evaluating}
Binsheng Zhao, Leonard~P James, Chaya~S Moskowitz, Pingzhen Guo, Michelle~S Ginsberg, Robert~A Lefkowitz, Yilin Qin, Gregory~J Riely, Mark~G Kris, and Lawrence~H Schwartz.
\newblock {Evaluating variability in tumor measurements from same-day repeat CT scans of patients with non--small cell lung cancer}.
\newblock \emph{Radiology}, 252\penalty0 (1):\penalty0 263--272, 2009.

\bibitem[Zheng and Yang(2021)]{zheng2021rectifying}
Zhedong Zheng and Yi~Yang.
\newblock {Rectifying pseudo label learning via uncertainty estimation for domain adaptive semantic segmentation}.
\newblock \emph{International Journal of Computer Vision}, 129\penalty0 (4):\penalty0 1106--1120, 2021.

\bibitem[Zhou et~al.(2021)Zhou, Sodha, Pang, Gotway, and Liang]{zhou2021models}
Zongwei Zhou, Vatsal Sodha, Jiaxuan Pang, Michael~B Gotway, and Jianming Liang.
\newblock Models genesis.
\newblock \emph{Medical Image Analysis}, 67:\penalty0 101840, 2021.

\end{thebibliography}
\end{document}